\documentclass{article}

\usepackage[utf8]{inputenc}
\usepackage{multicol}
\usepackage{multirow}
\usepackage[numbers]{natbib}
\usepackage{graphicx}
\usepackage{siunitx}
\usepackage{subcaption}
\usepackage{todonotes}
\usepackage{url}
\usepackage{xspace}

\usepackage{amsmath,amssymb,pifont}
\usepackage{algorithm,algorithmic}
\usepackage{amstext,amssymb,amsmath}
\usepackage{amsthm}
\usepackage{comment}
\usepackage{booktabs}
\usepackage{times}
\usepackage{lipsum}
\usepackage[shortlabels]{enumitem}
\usepackage{wrapfig}
\usepackage{array}
\usepackage{csvsimple}
\usepackage[multidot]{grffile}
\usepackage{bbm}
\usepackage{dblfloatfix}
\usepackage{makecell}
\usepackage{bbm, dsfont}
\usepackage{hyperref}
\usepackage{bbold}
\usepackage{mathtools}
\usepackage{xcolor}
\usepackage{hyperref}

\def\cA{\mathcal{A}}
\def\cD{\mathcal{D}}

\def\cS{\mathcal{S}}

\renewcommand{\Pr}{\mathop{\mathbf{Pr}}}

\def\calN{\mathcal{N}}

\newtheorem{lem}{Lemma}[section]

\newtheorem{defn}[lem]{Definition}

\renewcommand{\epsilon}{\varepsilon}

\title{Toward Training at ImageNet Scale with Differential Privacy}

\date{February 8, 2022}

\author{
Alexey Kurakin$^{1}$,
Shuang Song$^{1}$,
Steve Chien$^{1}$,
Roxana Geambasu$^{1,2}$,\\
Andreas Terzis$^{1}$,
Abhradeep Thakurta$^{1}$\\
$^{1}$ Google Research, $^{2}$ Columbia University
}

\begin{document}

\maketitle

\section*{Abstract}

Differential privacy (DP) is the de facto standard for training machine learning (ML) models, including neural networks, while ensuring the privacy of individual examples in the training set.
Despite a rich literature on how to train ML models with differential privacy,
it remains extremely challenging to train real-life, large neural networks with both reasonable accuracy and privacy.

We set out to investigate how to do this, using ImageNet image classification as a poster example of an ML task that is very challenging to resolve accurately with DP right now.
This paper shares initial lessons from our effort, in the hope that it will inspire and inform other researchers to explore DP training at scale.
We show approaches that help make DP training faster, as well as model types and settings of the training process that tend to work better in the DP setting.
Combined, the methods we discuss let us train a Resnet-18 with DP to $47.9\%$ accuracy and privacy parameters $\epsilon = 10, \delta = 10^{-6}$. This is a significant improvement over ``naive'' DP training of ImageNet models, but a far cry from the $75\%$ accuracy that can be obtained by the same network without privacy. The model we use was pretrained on the Places365 data set as a starting point.
We share our code at \url{https://github.com/google-research/dp-imagenet}, calling for others to build upon this new baseline to further improve DP at scale.

\section{Introduction}

Machine learning (ML) models are becoming increasingly valuable for improved performance across a variety of consumer products, from recommendations to automatic image classification and labeling.  However, despite aggregating large amounts of data, it is possible for models to encode — and thus, in theory, ``reveal'' — characteristics of individual entries from the training set.  For example, experiments in controlled settings have shown that language models trained using email datasets may sometimes encode sensitive information included in the training data~\cite{carlini2019secret} and may have the potential to reveal the presence of a particular user’s data in the training set~\cite{shokri2017membership}. As such, it is important to prevent the encoding of such characteristics from individual training entries.

The standard rigorous solution to this problem is differential privacy (DP)~\cite{DMNS,ODO}.
DP randomizes a computation over a dataset (such as training of an ML model) to bound the ``leakage'' of individual entries in the dataset through the output of the computation (the model).
The randomness ensures that the model is almost as likely to be output independent of the presence or absence of any individual entry in the training set.
This ``almost as likely'' is quantified by a privacy parameter, $\epsilon>0$.
If $\epsilon$ is small, the model cannot encode -- and thus cannot ``reveal'' -- much information about any individual entry.

Despite significant amounts of research in DP and machine learning, both on the theoretical front~\citep{chaudhuri2011differentially,kifer2012private,BST14,song2013stochastic,bassily2019private,mcmahan2017learning, WLKCJN17, pichapati2019adaclip,TAB19,feldman2019private,bassily2020stability,song2020characterizing} and on the empirical front~\citep{iyengar2019towards,abadi2016dpsgd,tramer2021dpfeatures} over the past decade, training ML models with DP remains challenging in practice, which limits its adoption.
First, DP often impacts utility, such as model accuracy.  The impact on accuracy comes from the randomness introduced into the computation to eliminate memorization of individual entries in some models, which can help accuracy.  Sometimes, the utility loss resulting from DP training is acceptable (or even could represents an improvement), but more often it is dramatically negative, making the resulting model useless.  For example, when you simply switch from SGD to DP-SGD~\cite{song2013stochastic,BST14,abadi2016dpsgd} to train a Resnet-18 or Resnet-50 on ImageNet, you get near-zero accuracy for any reasonable value of $\epsilon$.
Second, existing implementations of DP-SGD are inefficient~\cite{subramani2020fastdpsgd}.  Indeed, DP-SGD involves some expensive operations, like computing per-example gradients instead of per-minibatch gradients as non-private SGD.  This invalidates certain optimizations that are present in typical ML frameworks.  In production, this overhead can be prohibitive, especially if the model needs to be retrained often.  These two challenges have been preventing, on one hand, industry from adopting DP more widely and on the other hand, the research community from making progress on readying DP for wider adoption.  For example, most DP research papers evaluate DP algorithms on very small datasets, such as MNIST, CIFAR-10, or UCI.

This paper shares initial results from our ongoing effort to train a large image classification model on ImageNet using differential privacy while maintaining high accuracy and minimizing computational cost.
We split the paper in two parts: (1) the main body, which communicates the main lessons we have learned to date and (2) a hefty appendix that includes significantly more data from our experimentation.
Our goal is to inspire and inform other researchers who want to explore DP training at scale, including settings that work (our lessons) and those that may not (also covered in the appendix).

A first lesson we communicate (Section~\ref{sec:fast_dp}) is that a substantial amount of exploration -- of architectures, techniques, and hyperparameters -- is needed to discover a training setting that performs well with DP, and this exploration is impaired by the significant overheads of DP SGD implementations in most ML frameworks.
We recommend JAX as a good ML framework to perform such exploration in, because it is surprisingly effective at automatically optimizing otherwise very expensive DP operations.
A second lesson comes from our initial exploration of a few training settings for ImageNet, which combined let us DP-train a Resnet-18 to $47.9\%$ accuracy and privacy parameter $\epsilon = 10$ (Section~\ref{sec:imagenet_dp_result}), when pre-trained on Places365 data set.
This marks a good improvement compared to ``naive'' DP training, which achieves few percent accuracy for the same privacy parameter, but it remains far from the $75\%$ accuracy that can be obtained by the same network without privacy.

We share our code at \url{https://github.com/google-research/dp-imagenet}.
By sharing an early snapshot of the lessons, results, and code from our ongoing project, we hope to energize others to work on improving DP for {\em ambitious tasks} such as ImageNet, as a proxy for challenging production-scale tasks.
Our $47.9\%$-accuracy/$\epsilon = 10$ model should serve as a {\em baseline} to further improve upon to make DP training closer to practical at scale.

\section{Background}
\label{sec:background}

\subsection{Privacy Notions and DP-SGD}

\paragraph{ML Privacy Attacks.}
ML models are statistical aggregates of their underlying training data. Despite this, they have been shown to contain -- and therefore, in theory, expose to attacks -- information about the specific examples used to train them.
For instance trained ML models, and even their predictions, have been shown to enable membership inference attacks \cite{backes2016membership,dwork2015robustTraceability,shokri2017membership} (e.g., an learns that a particular user was in the training set for a disease detection model), and reconstruction attacks \cite{carlini2019secret,dinurNissim2003revealing,dwork2017exposed,carlini2020extracting} (e.g., an attacker reconstructs social security numbers from a language model).

\paragraph{Differential Privacy (DP).}
DP randomizes a computation over a dataset (such as training of an ML model) to bound the ``leakage'' of individual entries in the dataset through the output of the computation (the model).  
Intuitively, enforcing DP on the training procedure for an ML model ensures that the model is almost as likely to be output independent of the presence or absence of any individual entry in the training set; hence, the model cannot encode, and thus leak, much information about any individual entry.
DP is known to address the preceding ML privacy attacks~\cite{shokri2017membership,dwork2017exposed,carlini2019secret,Bargav2019evaluating}.
At a high level, membership and reconstruction attacks work by finding data
points that make the observed model more likely: if those points were in the
training set, the likelihood of the observed output increases.  DP prevents
these attacks, as no specific data point can drastically increase
the likelihood of the model outputted by the training procedure.

We state the formal definition of DP here.
\begin{defn}[Differential privacy \citep{DMNS, ODO}] A randomized algorithm $\cA$ is $(\epsilon,\delta)$-differentially private if, for any pair of datasets $D$ and $D'$ differing in exactly one data point (called {\em neighboring datasets}) i.e., one data point is added or removed, and for all events $\cS$ in the output range of $\cA$, we have 
$$\Pr[\cA(D)\in \cS] \leq e^{\epsilon} \cdot \Pr[\cA(D')\in \cS] +\delta,$$
where the probability is over the randomness of $\cA$. 
\label{def:diiffP}
\end{defn}
For meaningful privacy guarantees, $\epsilon$ is assumed to be a small constant, and $\delta \ll 1/n$ where $n$ is the size of $D$.

\paragraph{DP-SGD.}
DP-SGD is currently the most widely-used differentially private machine learning algorithm in practice. Along with its practical success, it also provides optimal privacy/utility trade-offs analytically~\cite{BST14,bassily2019private}. 
The full algorithm is described in Algorithm~\ref{alg:dpsgd}.

\begin{algorithm}[ht]
\caption{Differentially private stochastic gradient descent (DP-SGD)}
\begin{algorithmic}[1]
\REQUIRE Data set $D=\{d_1,\cdots,d_n\}$ with $d_i\in \cD$, loss function: $\ell:\mathbb{R}^p\times\cD\to\mathbb{R}$, clipping norm: $C$, 
number of iterations: $T$, noise multiplier: $\sigma$
\STATE Randomly initialize $\theta_0$.
\FOR{$t = 0,\dots,T-1$}
\STATE Randomly select a mini-batch of examples $B_t \subseteq D$
{\STATE {$g_t \leftarrow \sum_{d \in B_t}{\sf clip}\left(\nabla \ell(\theta_t;d)\right)$}, where {${\sf clip}(v)=v\cdot\min\left\{1,\frac{C}{\|v\|_2}\right\}$}.\label{step:clip}}
{\STATE $\theta_{t+1} \leftarrow$ {one step of first order optimization with gradient $g_t+\calN\left(0,(\sigma C)^2\right)$}\label{step:noiseDPSGD}}
\ENDFOR
{\STATE {\bf return} $\frac{1}{T}\sum\limits_{t=1}^T\theta_t$ or $\theta_T$\label{eq:lastDPSGD}.}
\end{algorithmic}
\label{alg:dpsgd}
\end{algorithm}


The privacy guarantee of DP-SGD comes from that of Gaussian mechanism with privacy amplification by subsampling and privacy composition. 
In Step~\ref{step:clip}, clipping upper-bounds the $\ell_2$ norm of each gradient in the mini-batch to be $C$, and thus the $\ell_2$ norm of $g_t$ changes by at most $C$ when we add or remove one example from the mini-batch $B_t$. Suppose $\sigma_t$ is the noise multiplier needed to achieve $\epsilon_{t}$-differential privacy in step $t$ when $B_t$ is arbitrarily chosen. 
If, instead, $B_t$ is formed by sampling $k$ examples \emph{u.a.r. and i.i.d.} from $D$, then privacy amplification by sampling~\cite{KLNRS,BST14,abadi2016dpsgd,pmlr-v89-wang19b} allows one to scale down the noise to $\sigma_t\cdot(k/n)$ while achieving the same privacy guarantee.
Similar subsampling schemes and analyses can be found in~\cite{soda-shuffling,feldman2020hiding,mcmahan2017learning,zhu2019poission}.
We can then accumulate the privacy loss in each step using composition, and obtain the final privacy cost of releasing all intermediate models $\{\theta_t\}_{t=1}^T$ as $\epsilon = O\left(\sqrt{T} \epsilon_t\right)$.

In this paper we use DP-SGD privacy analysis to compute $(\epsilon, \delta)$-DP guarantees 
for our experiments.
We report $\epsilon$ at $\delta = 10^{-6} \approx \frac{1}{\textrm{DATASET SIZE}}$
in most of the results.
However using smaller $\delta$ would only result in minor increase of $\epsilon$,
see Appendix~\ref{app:eps_detla_tradeoff}.

\subsection{Automatically Fast DP Training with JAX}\label{sec:fast_dp}

The previous section describes how DP-SGD works at a high level.
To understand why it is slow, let us compare SGD with DP-SGD in more detail.
One step of SGD works as follows: (0) Draw a batch of examples from the dataset. (1) ({\em Forward Pass}) Compute the loss function on the batch. (2) ({\em Backward Pass}) Then compute an average of the gradients with respect to the model parameters, and apply the average gradients to the model parameters using a chosen learning rate and optimizer.

DP-SGD differs as follows.  In each step, gradients are computed individually for each example instead of for the whole batch. Then, these per-example gradients are clipped to be within an $\ell_2$-norm, $C$, to control the sensitivity of the gradient average computation. After clipping, gradients of all examples are added together and Gaussian noise is added to the gradient vector, per the Gaussian mechanism described above. Finally, this clipped and noised gradient vector is applied to model parameters using the chosen learning rate and optimizer.

This per-example gradient clipping slows down DP-SGD compared to SGD for the following reason.
By default, ML frameworks -- such as Tensorflow and PyTorch -- do not directly compute per-example gradients and only provide aggregate, per-batch gradients. A na\"ive way to obtain per-example gradients is to loop over the examples in the batch, or to configure a batch size equal to one example. Unfortunately, this approach loses all the benefits of parallel batch computations and results in $N$ times slowdown compared to regular SGD, where $N$ is the batch size.

However, the lack of parallelization is not an inherent issue of per-example clipping.
If one manually writes the mathematical expressions for per-example gradient clipping, then it can be done in a parallel way at a cost of approximately one Forward and two Backward Passes. The Forward Pass proceeds as usual. At a first Backward Pass, norms of per-example gradients are accumulated. At a second Backward Pass, normalized gradients are computed. Thus, beyond this ``1.5x'' overhead, the overhead of a DP-SGD instantiation within a specific ML framework is probably caused by a suboptimal implementation of these operations inside the ML framework, rather than some inherent inefficiency of DP-SGD. This idea is inspired by~\cite{goodfellow2015efficient}.

Some DP-SGD libraries~\cite{opacus2021} take the approach of manually implementing necessary per-example operations in an efficient way.
This is valuable, however it can be time-consuming, error prone, and, importantly, difficult to evolve as new techniques for more effective DP-SGD training arise that may require changes within the underlying mathematical formulas.
Thus we prefer to avoid such manual approach.

\cite{subramani2020fastdpsgd} first observed that using JAX, a high-performance computational library based on XLA one can do efficient auto-vectorization and just-in-time compilation of the mathematical expressions needed for evaluating the clipped mini-batch gradient. In this work too, we rely on JAX's autovectorization capabilities to compute the clipped minibatch gradients necessary for the execution of DP-SGD. While~\cite{subramani2020fastdpsgd} empirically demonstrated the observation on CIFAR-10, we do the same on ImageNet.
We confirm that JAX can do all the parallelization and optimization necessary for per-example gradient computations {\em automatically}.

\begin{table}[t]
\centering
\begin{tabular}{ |c|c|c|c|c| } 
 \hline
 {\bf Dataset} & {\bf DP/No DP} & {\bf JAX (ours)} & {\bf Opacus} & {\bf TF-Privacy} \\ 
 \hline
 MNIST & DP & 1.48 & 1.95 & 6.35 \\
 MNIST & No DP & 0.68 & 0.52 & 0.96 \\
 \hline
 CIFAR10 & DP & 1.47 & 2.64 & 11.80 \\
 CIFAR10 & No DP & 0.39 & 0.61 & 1.14 \\
 \hline
\end{tabular}
\caption{Average time per training epoch in seconds, for different DP-SGD implementations. All experiments use a single V100 GPU.}
\label{table:perf_mnist_cifar}
\end{table}

\begin{table}[t]
\centering
\begin{tabular}{ |c|c|c|c| } 
 \hline
\multirow{2}{*}{\bf Dataset} & \multirow{2}{*}{\bf DP/No DP} & \multicolumn{2}{|c|}{\bf JAX (ours)} \\
\cline{3-4}
          &    & {\bf Resnet-18} & {\bf Resnet-50} \\
 \hline
 ImageNet & DP & 555.05 & 546.69 \\
 ImageNet & No DP & 275.5 & 365.96 \\
 \hline
\end{tabular}
\caption{Average time per training epoch in seconds for ImageNet on JAX. Experiments run on eight V100 GPUs.
It could be seen that DP-SGD training time for Resnet-18 and Resnet-50 is similar
despite the fact that Resnet-18 is a smaller model.
We didn't investigate this performance difference in details,
but suspect that it might be related to the fact that Resnet-50 using bottleneck residual blocks, while Resnet-18 is using non-bottleneck blocks.
}
\label{table:perf_imagenet}
\end{table}

Table~\ref{table:perf_mnist_cifar} shows the performance of DP and non-private training on MNIST and CIFAR10 (corroborating the observation by~\cite{subramani2020fastdpsgd}).
For both datasets we are using small convnets which are commonly used in DP literature~\cite{abadi2016dpsgd,papernot2020tempered}.
Our results measure the time per training epoch in seconds, averaged over entire training (15 epochs for MNIST, 90 epochs for CIFAR10).
To minimize influence of data loader on the performance we cached datasets in memory.
For non-private training, JAX and PyTorch do not show a consistent advantage over one another in these experiments, although they both show an advantage over Tensorflow.
For DP training, JAX is consistently faster than both Opacus and TF-Privacy: $24-44\%$ improvement per epoch over Opacus and $76-86\%$ over TF-Privacy.

These performance improvements can be quite important for research explorations with large networks, such as Resnet-18/Resnet-50 on ImageNet, whose runtimes without privacy are already orders of magnitude higher.
Table~\ref{table:perf_imagenet} shows the runtime per training epoch in seconds for ImageNet on JAX when run on eight GPUs in parallel, for Resnet-18 and Resnet-50.
Compared to non-private training, our JAX implementation is within 2x overhead, which we deem is fairly close to that rough theoretical best of ``1.5x,'' and also within reasonable realm for DP exploration.

\section{Effective Training Settings for DP}\label{sec:imagenet_dp_result}

Armed with this relatively faster JAX engine for DP-SGD training, we set out to answer the following question: {\em Of the multitude of potential training settings -- including model architectures, hyperparameter values, batch sizes, known DP methods, with their own settings -- which combination can lead to both good accuracy and privacy on ImageNet?}
We are still at the beginnings of answering this question, and many more explorations remain to be done, but in this section we summarize a few observations we have made so far.
They are:

\begin{enumerate}
    \item[{\bf Obs.~1:}] Choice of model matters, in particular smaller models tend to work better.
    \item[{\bf Obs.~2:}] More epochs is better than lower noise.
    \item[{\bf Obs.~3:}] Hyperparameter tuning makes a big difference.
    \item[{\bf Obs.~4:}] Extremely large batch size improves the privacy-utility tradeoff.
    \item[{\bf Obs.~5:}] Transfer learning from public data significantly boosts accuracy.
\end{enumerate}

Similar observations can be found scattered in prior DP literature, but to our knowledge they have not been all combined and evaluated together.
Our contribution thus lies in evaluating these methods on the ambitious ImageNet classification task, where we find they yield a reasonable new baseline for DP-training at scale.

\subsection{Obs.~1: Choice of model matters}

A large variety of models have been developed for non-private training on ImageNet and the question of which model design (if any) is more suitable for DP training is still wide open.
In our exploration, we wanted to start with not too small but also not too big models, so we chose the Resnet-v2~\cite{resnetv2} model family.
Within that, we chose to focus on Resnet-50 and Resnet-18: the former is generally considered as a sufficiently powerful model for this dataset; the latter is a smaller model for comparison.
We applied known rules to adapt these models for DP training.
In particular, any cross-batch procedure during training -- such as through batch normalization -- breaks privacy analysis of DP-SGD and must be replaced with a per-batch procedure.  
In Resnet-50 and Resnet-18, we thus replaced batch normalization with group normalization~\cite{groupnorm2018}.

In evaluating Resnet-50 and Resnet-18, we observe that model size impacts accuracy differently than one might expect in non-private training.
In non-private training, one rule of thumb is that complex learning tasks, such as image classification on ImageNet, tend to do better with larger, more expressive networks.
In DP training, larger is not always better.
To obtain a desired level of privacy, $\epsilon$, DP-SGD calls for adding a fixed amount of noise, $\sigma$, to every parameter {\em independently of the model size}.
Thus, applying DP-SGD to large models causes a larger shift in the parameters of large models compared to smaller ones, which can lead to a negative effect on accuracy.

\begin{table}[t]
\centering
\begin{tabular}{ |c|c|c|c|c|c| } 
 \hline
\multirow{2}{*}{\bf No DP}  & {\bf 10 epochs} & {\bf 10 epochs} & {\bf 30 epochs} & {\bf 60 epochs} & {\bf 90 epochs} \\ \cline{2-6}
  & {\bf tuned LR} & \multicolumn{4}{c|}{{\bf learning rate = 0.4}} \\ 
 \hline
 Resnet-18 & 58.7\% & 57.6\% & 67.5\% & 70.3\% & 70.8\% \\
 Resnet-50 & 62.1\% & 60.5\% & 72.0\% & 74.5\% & 75.3\% \\
 \hline
\end{tabular}
\caption{Comparison of Resnet-18 and Resnet-50 top-1 accuracy without DP, depending on number of training epochs.}
\label{table:imagenet_nodp}
\end{table}

\begin{table}[t]
\centering
\begin{tabular}{ |c|c|c|c|c|c|c|c|c|c| } 
 \hline
 \multirow{2}{*}{\bf DP} & \multicolumn{8}{|c|}{\bf privacy loss bound $\mathbf{\epsilon}$} \\
 \cline{2-9}
 & $\mathbf{4.6}$ & $\mathbf{13.2}$ & $\mathbf{71}$ & $\mathbf{\approx10^7}$ & $\mathbf{10^9}$ & $\mathbf{10^{11}}$ & $\mathbf{10^{13}}$ & $\mathbf{10^{15}}$ \\
 \hline
 Resnet-18 & 3.7\% & 6.9\% & 11.3\% & 45.7\% & 55.4\% & 56.0\% & 56.3\% & 56.4\% \\
 Resnet-50 & 2.4\% & 5.0\% & 7.7\%  & 44.3\% & 58.8\% & 57.8\% & 58.2\% & 58.6\% \\
 \hline
\end{tabular}
\caption{Comparison of the best Resnet-18 and Resnet-50 top-1 accuracies obtained at 10 epochs and batch size 1024, for various values of the privacy loss bound $\epsilon$ and fixed $\delta = 10^{-6}$.
The different $\epsilon$ values are obtained by applying to the gradient vector noise from Gaussian distributions with different standard deviations, $\sigma \in \{0.56, 0.42, 0.28, \num{2.8e-2}, \num{2.8e-3}, \num{2.8e-4}, \num{2.8e-5}, \num{2.8e-6}\}$, respectively.
Clipping norm is fixed $C = 1$.
Each accuracy number for Resnet-18 is obtained by sweeping over learning rates in $\{1, 2, 4, 8, 16, 40\}$.
Each accuracy number for Resnet-50 is obtained by sweeping over learning rates in $\{0.4, 0.8, 1.6, 4, 8, 16, 40\}$.
}
\label{table:resnet18_vs_50_naive_dp}
\end{table}

Table~\ref{table:imagenet_nodp} compares accuracies of Resnet-18 and Resnet-50 without DP, while Table~\ref{table:resnet18_vs_50_naive_dp} compares them with DP.
Without DP (Table~\ref{table:imagenet_nodp}), we show accuracies achieved at different epochs, with tuned learning rate (LR) and a fixed $LR=0.4$.
The larger Resnet-50 tends to do better across the board.
With DP (Table~\ref{table:resnet18_vs_50_naive_dp}), we show accuracies achieved at 10 epochs for various $\epsilon$ values with batch size 1024.
Resnet-18 tends to outperform Resnet-50 when $\epsilon$ is low (and hence the privacy guarantees are better).
On the other hand, when $\epsilon$ is very large, Resnet-50 starts to outperform Resnet-18 again, consistent with the non-private behavior.
Further research is needed into whether simply using a smaller network, as we do here, is the right way to reduce dimensionality and the impact of DP noise on NN training.
Another option might be to adapt $\sigma$ to the statistics of individual parameters, as done in AdaClip~\cite{pichapati2019adaclip}.
Overall, though, our results point to the importance of adapting model architecture to DP training, since what tends to work for non-private training may not work for DP training.

We also investigated the effect of model structure and size in CIFAR-10.  Appendix~\ref{app:cifar10} describes the methodology and results.
There too, in the low $\epsilon$ regime, smaller and simpler networks tend to have better accuracy than networks that do best in non-private learning.

There is some limited prior work studying relationship between model size and 
accuracy with DP training.
We found that such results for image models (Figure 1 in~\cite{papernot2020making}, Table 20 in~\cite{tramer2021dpfeatures}) agree with ours~--- larger models tend to perform worse.
However results for language models are quite opposite.
As reported in~\cite{li2022llmdp} during DP fine-tuning of language models,
larger model architecture tends to be beneficial.
We hypothesize that this is related to the following facts.
First of all, the gradient spectrum of language models differ significantly from that of vision models~\cite{pmlr-v97-agarwal19b}.
Second, (noisy) gradient descent tends to respect low-dimensional subspace induced by the gradients~\cite{song21clipping}.
Thus finetuning of language models might be happening in lower dimensional subspace
when larger model size is beneficial.
Nevertheless, future research in this direction is needed to fully understand these differences.

\subsection{Obs.~2: More epochs is better than lower noise}

Privacy loss bound $\epsilon$ of a DP-SGD training depends on noise multiplier $\sigma$ and number of training steps.
Higher noise multiplier $\sigma$ leads to lower $\epsilon$, but typically worse accuracy.
Longer training leads to higher $\epsilon$ as described in Section~\ref{sec:background}.
At the same time, at least in non-private setting, training for longer typically helps to achieve higher accuracy.
Thus if one to set a fixed $\epsilon$ and wants to maximize accuracy, a natural question arises:
{\em Is it better to train longer with higher noise multiplier or train for fewer epochs with lower noise?}


\begin{figure}[t]
\centering
\begin{subfigure}{.33\textwidth}
  \centering
  \includegraphics[width=\linewidth]{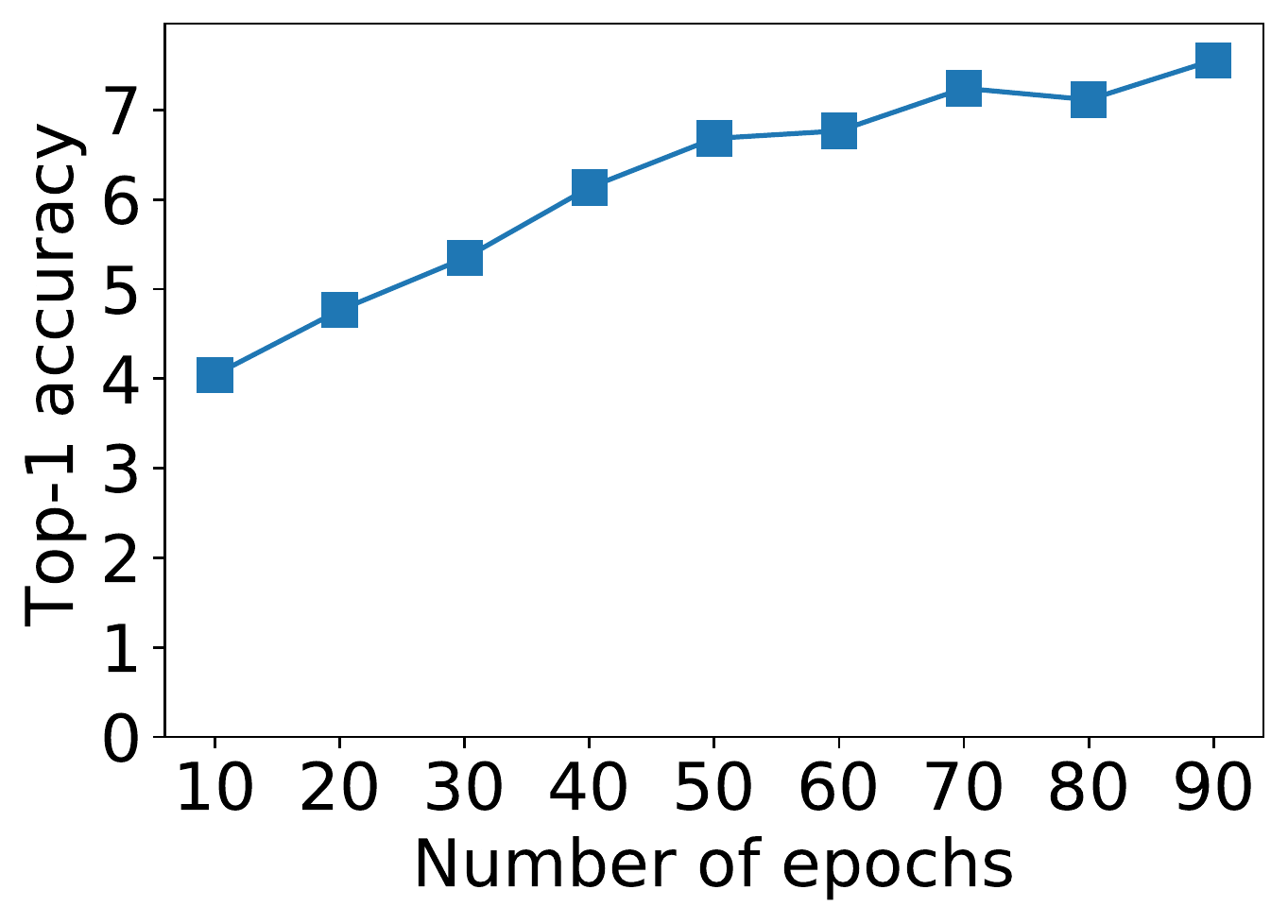}
  \caption{$\epsilon = 4.57$}
\end{subfigure}%
\begin{subfigure}{.33\textwidth}
  \centering
  \includegraphics[width=\linewidth]{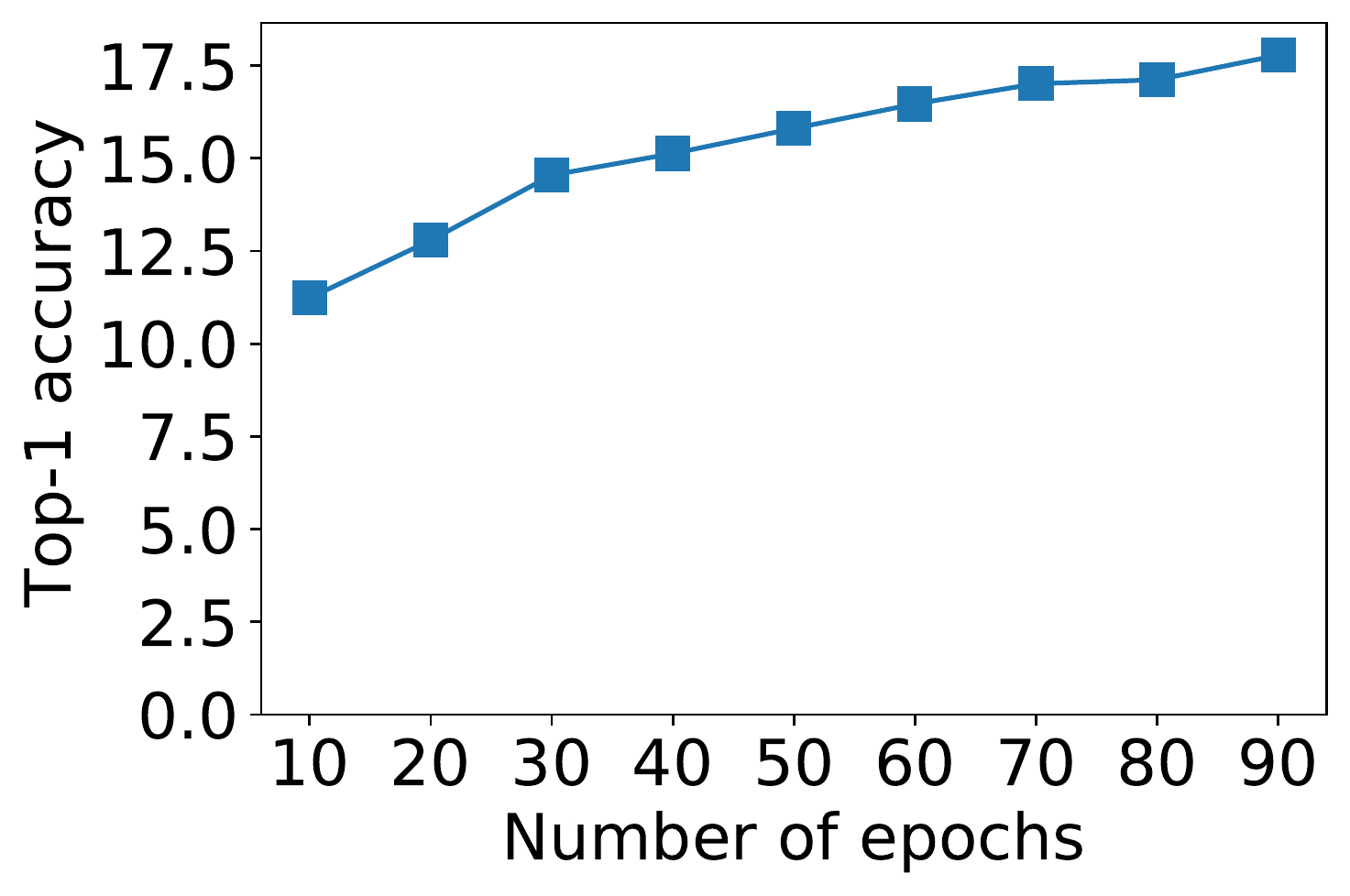}
  \caption{$\epsilon = 71.5$}
\end{subfigure}%
\begin{subfigure}{.33\textwidth}
  \centering
  \includegraphics[width=\linewidth]{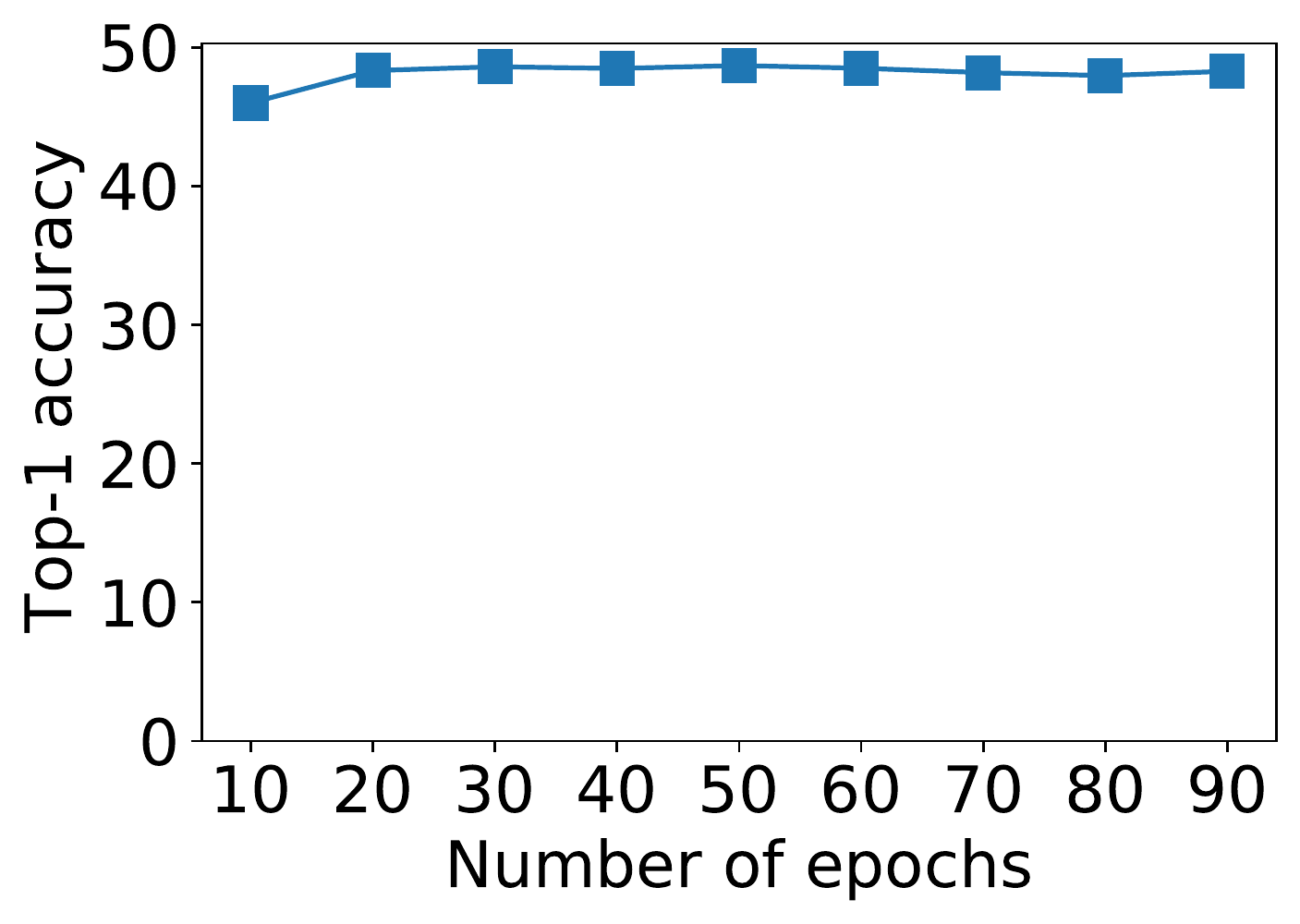}
  \caption{$\epsilon = 9.7 \cdot 10^6$}
\end{subfigure}%
\caption{
Resnet-18 accuracy while training for different number of epochs but keeping constant $\epsilon$.
We tuned learning rate independently and reported the best accuracy for each combination of $\epsilon$ and number of epochs.
Specific accuracy numbers for these plots are available
in Appendix~\ref{app:fixed_eps_exp}.}
\label{fig:epochs_fixed_eps}
\end{figure}

Figure~\ref{fig:epochs_fixed_eps} shows the accuracy achieved by Resnet-18 when training for increasing number of epochs while keeping the privacy loss $\epsilon$ constant by correspondingly decreasing the noise multiplier $\sigma$.
In these experiments, training longer and with higher noise multiplier gives better accuracy than training for fewer epochs but with less noise.
Nevertheless, accuracy reaches an upper bound somewhere between 40 and 70 epochs.

\subsection{Obs.~3: Hyperparameter tuning makes a big difference}

Training with DP-SGD requires setting extra hyper-parameters, such as the noise multiplier $\sigma$ and the clipping norm $C$.
Moreover, we observed that additional tuning of learning rate is typically necessary.
The noise multiplier $\sigma$ should be chosen to satisfy desired privacy budget.
The clipping norm and learning rate should be tuned jointly for the best result.
Appendix~\ref{app:c_lr_sweep} gives results from our exploration of various values for learning rate and clipping norm.
Here, we formulate a few observations we made experimentally on how to tune these parameters.

We observed that there is typically some threshold value $\tilde C$, such that the best private accuracy is obtained when the clipping norm is smaller than $\tilde C$.
Below the $\tilde C$ threshold, a wide range of values of the clipping norm can be used as long as the learning rate is adjusted accordingly.
Specifically, when the clipping norm is decreased $k$ times, the learning rate should be increased $k$ times to maintain similar accuracy.
However, a larger learning rate (with smaller clipping norm) may lead to less stable training, thus we recommend keeping the clipping norm close to $\tilde C$.

Overall, we recommend the following procedure to tune the clipping norm and learning rate:

\begin{enumerate}
    \item Find a good set of hyperparameters in a non-private case (for example from a public dataset).
      Let $\alpha_{pub}$ be the good non-private learning rate.
    \item Sweep over various values of the clipping norm $C$ with fixed learning rate $\alpha_{pub}$ and zero noise $\sigma = 0$. Find the smallest $\tilde C$ for which DP model accuracy remains close to the non-private model accuracy.
    \item Set the clipping norm to $\tilde C$ and set the noise multiplier $\sigma$ based on desired privacy budget.
    Run a learning rate sweep and find a good learning rate $\tilde \alpha$.
    Typically $\tilde C$ and $\tilde \alpha$ would be a reasonably good combination of hyperparameters for private training.
    \item Further grid search in the vicinity of $\tilde C$ and $\tilde \alpha$ may bring additional improvement of the accuracy.
\end{enumerate}

It is important to acknowledge that the preceding procedure for hyper-parameter tuning does {\em not} preserve privacy.
We leave it for future work to develop such a private procedure, perhaps based on recent results~\cite{papernot2021dphyperparam}.

\subsection{Obs.~4: Large batch size improves the privacy-utility tradeoff}

In theory, it is known that large minibatch size improves utility in DP-SGD, with full-batch training giving the best outcome~\cite{talwar2014largebatch,bassily2020stability}.  (Here, full-batch training refers to using the entire dataset as the batch for each iteration.)
Recently,~\cite{anil2021dpbert} corroborated this observation empirically on language models (e.g., BERT). Since, language models tend to have a different gradient profile (see Figure 5 in~\cite{pmlr-v97-agarwal19b}) than vision models, it is non-obvious at the outset that a similar phenomenon would hold in the problem setup we consider. Via empirical evaluation, we demonstrate that the observations of~\cite{anil2021dpbert} extend to vision models too.

In practice, for large datasets like ImageNet, full batch does not fit into GPU/TPU memory,
and an attempt to distribute it will result in a prohibitively large number of accelerators.
For example, Resnet-50 training on one accelerator will typically fit a batch size of 64 or 128.
Thus, full-batch training on ImageNet would require more than 9000 accelerators in distributed data-parallel regime.
Fortunately, large batch and full batch training can be simulated by accumulating gradients over several steps before applying them, a process called {\em virtual steps} and already implemented in some DP libraries, including Opacus.

\begin{table}[t]
\centering
\begin{tabular}{ |c|c|c|c|c| } 
 \hline
 {\bf Batch size} & {\bf 1024} & {\bf 4*1024} & {\bf 16*1024} & {\bf 64*1024} \\
 {\bf Num epochs} & {\bf 10} & {\bf 40} & {\bf 160} & {\bf 640} \\ 
 \hline
 {\bf Accuracy} & 56\% & 57.5\% & 57.9\% & 57.2\% \\
 {\bf Privacy loss bound $\mathbf{\epsilon}$} & $9.8\cdot 10^8$ & $6.1\cdot 10^7$ & $3.5\cdot 10^6$ & $6.7\cdot 10^4$ \\
\hline
\end{tabular}
\caption{Resnet-18 accuracy and privacy ($\epsilon$) on ImageNet with increasingly large batches. All experiments used noise multiplier $\sigma = 0.001 \cdot \sqrt{8} \cdot \frac{\textrm{BatchSize}}{1024}$ and learning rate $16$.}
\label{table:imagenet_batch_scaling}
\end{table}

\begin{figure}[ht]
\centering
\includegraphics[width=0.4\textwidth]{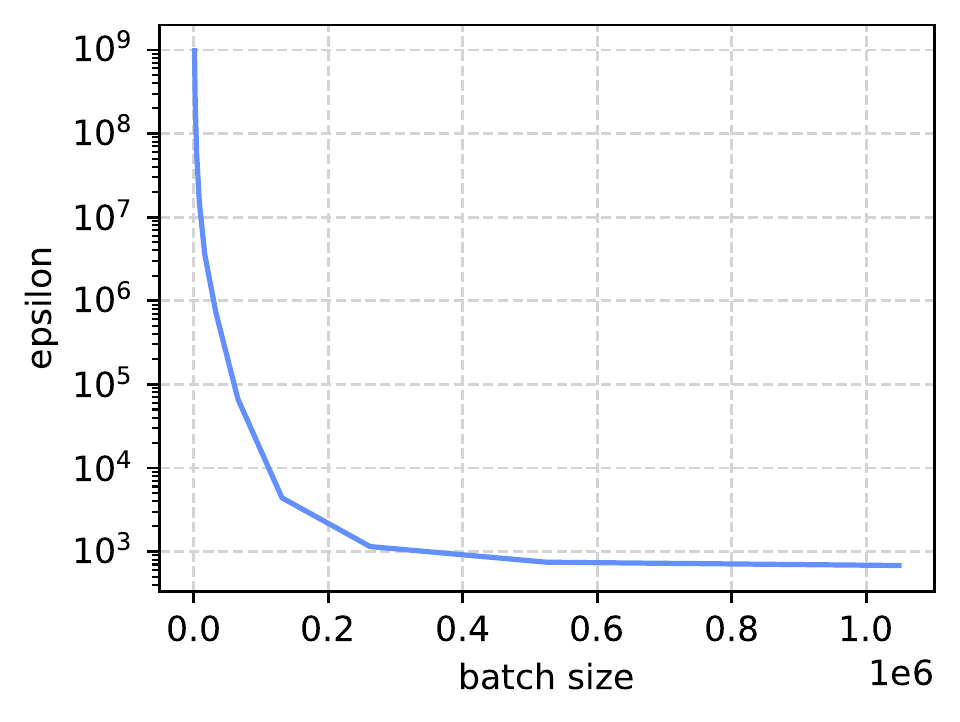}
\caption{$\epsilon$ value as batch size increases, with noise / batch size ratio and number of steps kept fixed.}
\label{fig:batch privacy}
\end{figure}

Table~\ref{table:imagenet_batch_scaling} shows that joint scaling of batch size, number of training epochs and noise multiplier
can lead to decrease of $\epsilon$ while maintaining a similar accuracy level.
A batch size of 16*1024 means to make one gradient step, we accumulate in memory 16 virtual steps each with 1024 examples.
As we increase the batch size, we must keep the number of training {\em steps} the same to maintain the same accuracy.
This results in larger number training {\em epochs} as we increase batch size.
Table~\ref{table:imagenet_batch_scaling} is demonstrational: the $\epsilon$ values shown there are entirely unacceptable for privacy.
In Figure~\ref{fig:batch privacy}, we plot the $\epsilon$ value at different batch size, when the noise / batch ratio and the number of steps are fixed.
However, the same effect of large-batch size being advantageous can be witnessed for more reasonable values of $\epsilon$, albeit with much lower accuracy outcomes.
For example, with a batch size of $16*1024$ and 10 epochs, the Resnet-18 will obtain an accuracy of 5.8\% for $\epsilon=2.2\cdot 10^5$.
But with a much larger batch size of $1024*1024$ (full-batch training) and a corresponding larger number of epochs of 640 (to preserve the total number of steps), the Resnet-18 will obtain a roughly equivalent 6.2\% accuracy but for a (perhaps) more reasonable $\epsilon=72$.

A recent work~\cite{chourasia2021differential} studies full-batch DP-SGD algorithm as an instantiation of Langevin Diffusion, and show a tighter privacy/utility trade-offs as compared to the standard optimization viewpoint~\cite{BST14}. While the impact of this approach is yet to be realized in practice (as the improvement only holds for smooth and strongly convex losses), it is an important research direction to explore.
For our paper, however, we remain in the realm of standard optimization view of DP-SGD, and instead combine large-batch training with transfer learning from public data, which offers a significant boost in accuracy.

\subsection{Obs.~5: Transfer learning from public data significantly boosts accuracy}

Pre-training on ``public’’ data followed by DP fine-tuning on private data has previously been shown to improve accuracy on other benchmarks~\cite{tramer2021dpfeatures,yu2021nlpdpfinetuning,li2022llmdp}.
We confirm the same effect on ImageNet.
A big question with transfer learning is what public data to use for a given task.
We were surprised to see that reasonable choices are quite effective.

\begin{table}[t]
\centering
\begin{tabular}{|c|c|c|c|c|}
\hline
 & \textbf{10 epochs} & \textbf{40 epochs} & \textbf{70 epochs} \\
\hline
From scratch & 12.0\% & 19.0\% & 20.6\% \\
\hline
No frozen layers & 29.2\% & 34.9\% & 37.7\% \\
1 frozen block group & 29.8\% & 35.1\% & 38.0\% \\
2 frozen block groups & 30.7\% & 36.0\% & 38.8\% \\
3 frozen block groups & 32.5\% & \textbf{38.9\%} & \textbf{40.7\%} \\
4 frozen block groups & \textbf{33.5\%} & 36.3\% & 36.9\% \\
\hline
\end{tabular}
\caption{Comparison of finetuning vs training from scratch with differential privacy.
First row is training from scratch, other rows are finetuning with different number of frozen layers.
Privacy budget was set to $\epsilon = 10$ in all experiments.
We used batch size $4*1024$ and
each accuracy number was obtained by sweeping learning rate in $\{0.016, 0.048, 0.16, 0.48, 1.6, 4.8, 16.0\}$.
\textbf{Bold} number highlights the best accuracy in each column.}
\label{table:finetuning_vs_scratch}
\end{table}

We pre-trained our models on Places365~\cite{zhou2017placesdataset}, another image classification dataset, before fine-tuning them with DP-SGD on ImageNet.
Places365 contains 1.8M images of various scenes with 365 labels describing the scene. 
This dataset has only images of landscapes and buildings, not of animals as ImageNet, so it is quite different.
Thus we consider the pair (Places365, ImageNet) as a reasonable proxy for real world setups of public and private datasets.

In our experiments, we first non-privately train Resnet-18 on Places365 to $54.96\%$ accuracy
(see Appendix~\ref{app:places365pretraining}).
Then we strip the last linear layer of the model and replace it with a randomly initialized one with the 1000 output classes of the ImageNet classification.
Finally, we try different schemes of fine-tuning this model on ImageNet with DP-SGD.

In a first set of experiments, we compared finetuning with training from scratch and 
explored whether keeping some layers frozen helps to increase accuracy, see table~\ref{table:finetuning_vs_scratch}.
Unsurprisingly, finetuning is always better than training from scratch.
These results also suggest that freezing more layers tend to be better.
It's possible that freezing less layers may work well when batch size and number of epochs are increased,
however we didn't explore any further increase of number of epochs with finetuning in this paper.

\begin{table}[t]
\centering
\begin{tabular}{|c|c|c|c|c|c|c|}
\hline
Frozen & Batch & & & & & \\
block  & size $\rightarrow$ & 4*1024 & 16*1024 & 64*1024 & 256*1024 & 1024*1024 \\ \cline{2-2}
groups & Number of & & & & & \\
       & epochs $\downarrow$ & & & & & \\
\hline
  & 10 & 32.5\% & \textbf{39.6\%} & 33.0\% & 18.6\% &  3.3\% \\
3 & 40 & 38.9\% & 44.0\% & \textbf{44.9\%} & 36.4\% & 17.0\% \\
  & 70 & 40.7\% & 45.0\% & \textbf{47.9\%} & 41.7\% &   18.4\% \\
\hline
  & 10 & 33.5\% & 36.1\% & \textbf{37.0\%} & 33.6\% & 23.1\% \\
4 & 40 & 36.3\% & 37.2\% & \textbf{37.8\%} & 37.0\% & 33.1\% \\
  & 70 & 36.9\% & 37.7\% & 38.0\% & \textbf{38.1\%} & 34.7\% \\
\hline
\end{tabular}
\caption{Accuracy of Resnet18 model which was pre-trained on Places365 and finetuned with DP-SGD on ImageNet.
Each accuracy number was obtained by running learning rate sweep.
Noise multiplier for each experiment was chosen in a way that in the end of the training $\epsilon = 10$.
\textbf{Bold} numbers highlight the best accuracy in each row,
i.e. the best accuracy when number of training epochs and frozen blocks is fixed.
}
\label{table:imagenet_dp_finetuning}
\end{table}

In the second set of experiments we restrict ourself to freezing most of the network layers,
and in this setup studied how batch size and number of epochs affect the final private accuracy,
see table~\ref{table:imagenet_dp_finetuning}.
These experiments reinforces our observation that longer training is generally better when $\epsilon$ is fixed.
At the same we see that increase of batch size only helps to a certain point, after which it is actually hurting accuracy.
As was mentioned earlier, increase of batch size generally should be done together with increasing number of epochs.
Thus we hypothesize that number of training epochs in table~\ref{table:imagenet_dp_finetuning} was not large enough
to see benefits of batch sizes $256*1024$ and $1024*1024$.

\subsection{Combining everything together}

Combining all experiments and observations from previous sections, we managed to train a Resnet-18 on ImageNet
to $47.9\%$ accuracy with privacy budget $\epsilon=10$ by using the following hyperparameters:

\begin{itemize}
    \item Start with a model pre-trained on Places365.
    \item Finetune this model on ImageNet with DP-SGD for $70$ epochs with batch size $64*1024$.
    \item Use cosine decay learning rate schedule with a warmup for $1$ epochs. Maximum learning rate is $7.68$.
    \item Use Nesterov momentum optimizer with standard decay $0.9$.
    \item Use weight decay loss term with a coefficient $10^{-4}$.
    \item Set clipping norm $C = 1$ and choose noise multiplier $\sigma$ to satisfy privacy budget.
\end{itemize}

\section{Conclusion}
In this work we make a first attempt to train a model on ImageNet dataset with differential privacy.
To achieve this we study how various techniques and training parameters can affect accuracy of DP-SGD training.
Combination of all of this findings enables us to train Resnet-18 on ImageNet to $47.9\%$ accuracy with 
private budget $\epsilon = 10$.
While it may look relatively low compared to typical Resnet accuracy obtained in non-private training,
one should keep in mind that ``naive'' Resnet training with DP-SGD on ImageNet will typically result in
either only few percent accuracy or very high epsilon (which means lack of privacy).

\bibliographystyle{plain}
\bibliography{references}

\appendix

\section{Resnet training without DP}\label{app:train_nodp}

We trained Resnet-18 and Resnet-50 models without differential privacy for various number of epochs,
as shown table~\ref{table:imagenet_nodp}.
We used the following training setup in these experiments:

\begin{itemize}
    \item Nesterov momentum optimizer with momentum $0.9$.
    \item Learning rate warmup for the first $5$ epochs to maximum learning rate, followed by cosine decay to zero.
    In most of the non-private experiments we used the same maximum learning rate $0.4$ which was originally tuned for 90 epochs Resnet-50 training.
    In some experiments we did additional tuning of the learning rate, which may result in a few percent accuracy boost.
    \item Weight decay loss term with coefficient $10^{-4}$. We tried different values of weight decay coefficient and found that $10^{-4}$ is the best one.
    \item Total batch size was $1024$ and training was done on 8 v100 GPUs or 8-core TPU.
\end{itemize}

\section{Resnet training with DP-SGD}\label{app:train_dp}

For DP-SGD training we generally used similar setup as for non-private training (see appendix~\ref{app:train_nodp}): Nesterov momentum optimizer, learning rate warmup followed by cosine decay, weight decay $10^{4}$ and training on 8 GPUs or 8 TPU cores.
The main differences is that we used DP-SGD version of the optimizer~\cite{abadi2016dpsgd}, which requires to set two extra parameters:
clipping norm $C$ and noise multiplier $\sigma$. Another difference is that we have to re-tune learning rate compared to non-private training.
Additionally when training for 10 epochs we typically used only one epoch to warmup learning rate.

It should be noted that in our implementation Gaussian noise is added independently per GPU or per TPU core.
This means that total noise added to the gradients has standard deviation of $\sigma = \sqrt{8}\sigma_{1}$ where $\sigma_{1}$
is a standard deviation of noise added to single replica.
That's why many tables in this report are showing $\sigma / \sqrt{8}$ which is equal to standard deviation of per-replica noise.

We note that the $\epsilon$ of DP-SGD is computed through R\'enyi differential privacy (RDP) analysis, and the conversion between RDP to DP can be affected by the choice of RDP orders. In this paper, we use a function~\cite{compute_dp_sgd_privacy_lib} in TF-Privacy to compute the DP $\epsilon$. Due to the choice of RDP orders there, $\epsilon$ might be overestimated, especially in the low privacy / large $\epsilon$ regime. 

\section{Experiments with fixed epsilon and different number of epochs}\label{app:fixed_eps_exp}

We did a series of experiments where we set $\epsilon$ to a desired value
and then vary number of training epochs and other parameters.
To run these experiments we need to compute noise multiplier $\sigma$
based on desired $\epsilon$, batch size, number of training epochs, etc...
However privacy accountants from DP-SGD libraries typically provide
a way to compute $\epsilon$ given noise multiplier,
but not the other way around.
To overcome this difficulty we used a routine from Tensorflow Privacy which
computes $\epsilon$ based on noise multiplier
and did a binary search with branching by geometric mean
to find $\sigma$ corresponding to the desired $\epsilon$.

Plots with results of the experiments with fixed $\epsilon$ and different
number of epochs are provided in the Figure~\ref{fig:epochs_fixed_eps}
in the main body of the paper.
Table~\ref{table:epochs_fixed_eps} contains detailed numerical results
of these experiments.

\begin{table}[ht]
\centering
\begin{tabular}{ |c|c|c|c| } 
 \hline
 \bf{Num epochs} & $\epsilon = 4.57$ & $\epsilon = 71.5$ & $\epsilon = 9.7\cdot 10^6$ \\
 \hline
 \bf{10/1} & $4.0\%$ & $11.2\%$ & $45.5\%$ \\
 \bf{10}   & $3.6\%$ & $10.5\%$ & $46.0\%$ \\
 \bf{20}   & $4.8\%$ & $12.8\%$ & $48.3\%$ \\
 \bf{30}   & $5.3\%$ & $14.5\%$ & $48.6\%$ \\
 \bf{40}   & $6.1\%$ & $15.1\%$ & $48.5\%$ \\
 \bf{50}   & $6.7\%$ & $15.8\%$ & $48.7\%$ \\
 \bf{60}   & $6.8\%$ & $16.5\%$ & $48.5\%$ \\
 \bf{70}   & $7.2\%$ & $17.0\%$ & $48.2\%$ \\
 \bf{80}   & $7.1\%$ & $17.1\%$ & $48.1\%$ \\
 \bf{90}   & $7.6\%$ & $17.8\%$ & $48.3\%$ \\
 \hline
\end{tabular}
\caption{Resnet-18 accuracy while training for different number of epochs but keeping constant $\epsilon$.
For each of the experiments we did a learning rate sweep in $\{0.1, 0.2, 0.5, 1, 2, 4, 8, 16, 40\}$ and reported the best accuracy.
Row \textbf{10/1} correspond to training for 10 epochs with 1 epoch of learning rate warmup.
Experiments in all other row used 5 epochs of learning rate warmup.
}
\label{table:epochs_fixed_eps}
\end{table}

\section{Hyperparameter tuning for DP-SGD}\label{app:c_lr_sweep}

While longer training can generally help increase utility,
we found that training for 10 epochs is generally provide a reasonable idea of what we can expect in terms of accuracy.
Thus to be able to do large hyperparameter sweeps within limited compute budget
we restricted most of the hyperparameter sweeps to 10 epochs.

For 10 epoch training of Resnet-18
we did a joint sweep of learning rate and clipping norm for various value of noise multiplier $\sigma$.
For each value of $\sigma$ we did one fine grained sweep of learning rate from $\{1, 2, 4, 8, 16, 40\}$
and another coarse sweep where learning rate was increasing as a power of $10$.

\begin{figure}[ht]
\centering
\includegraphics[width=0.49\textwidth]{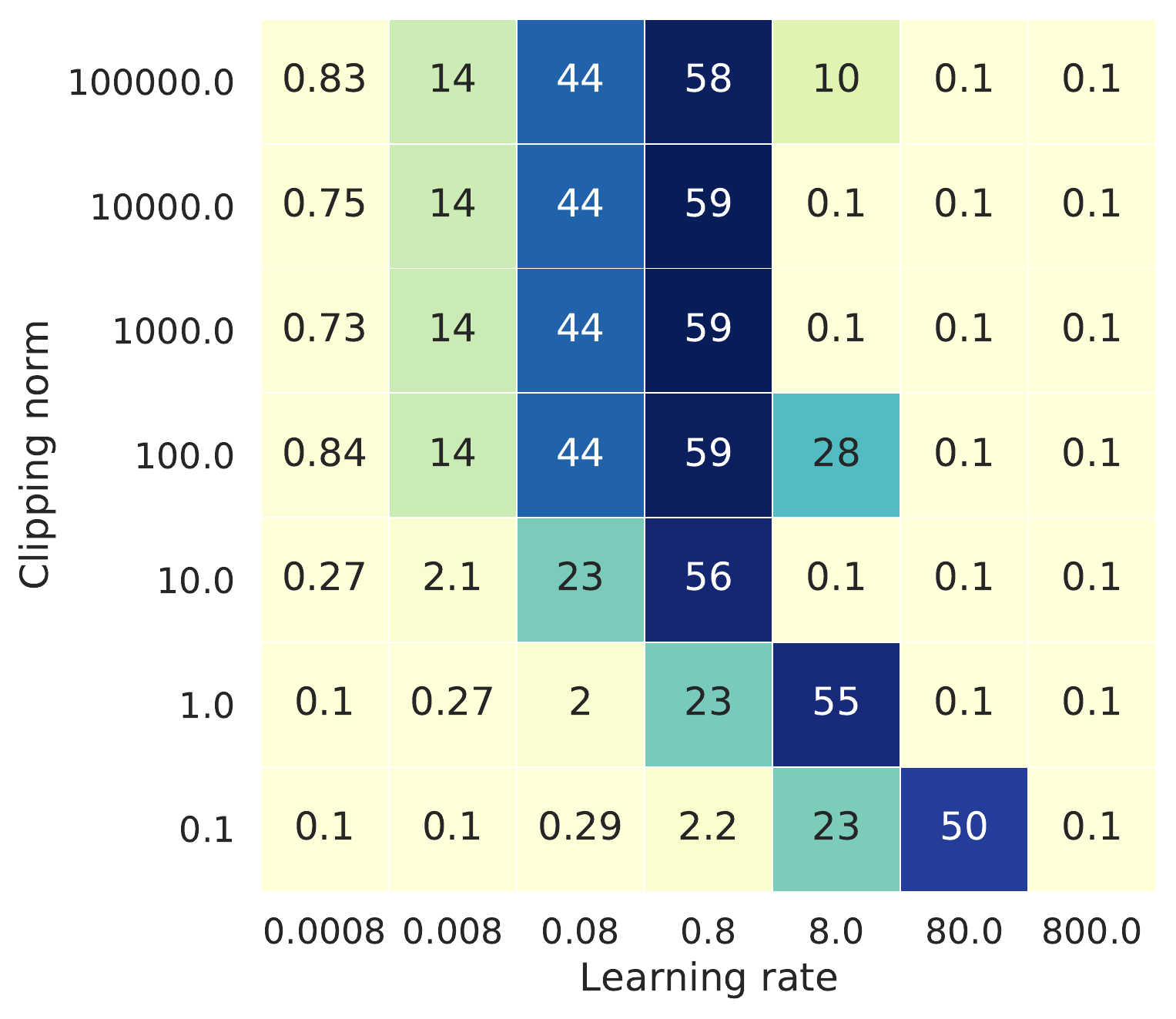}
\includegraphics[width=0.49\textwidth]{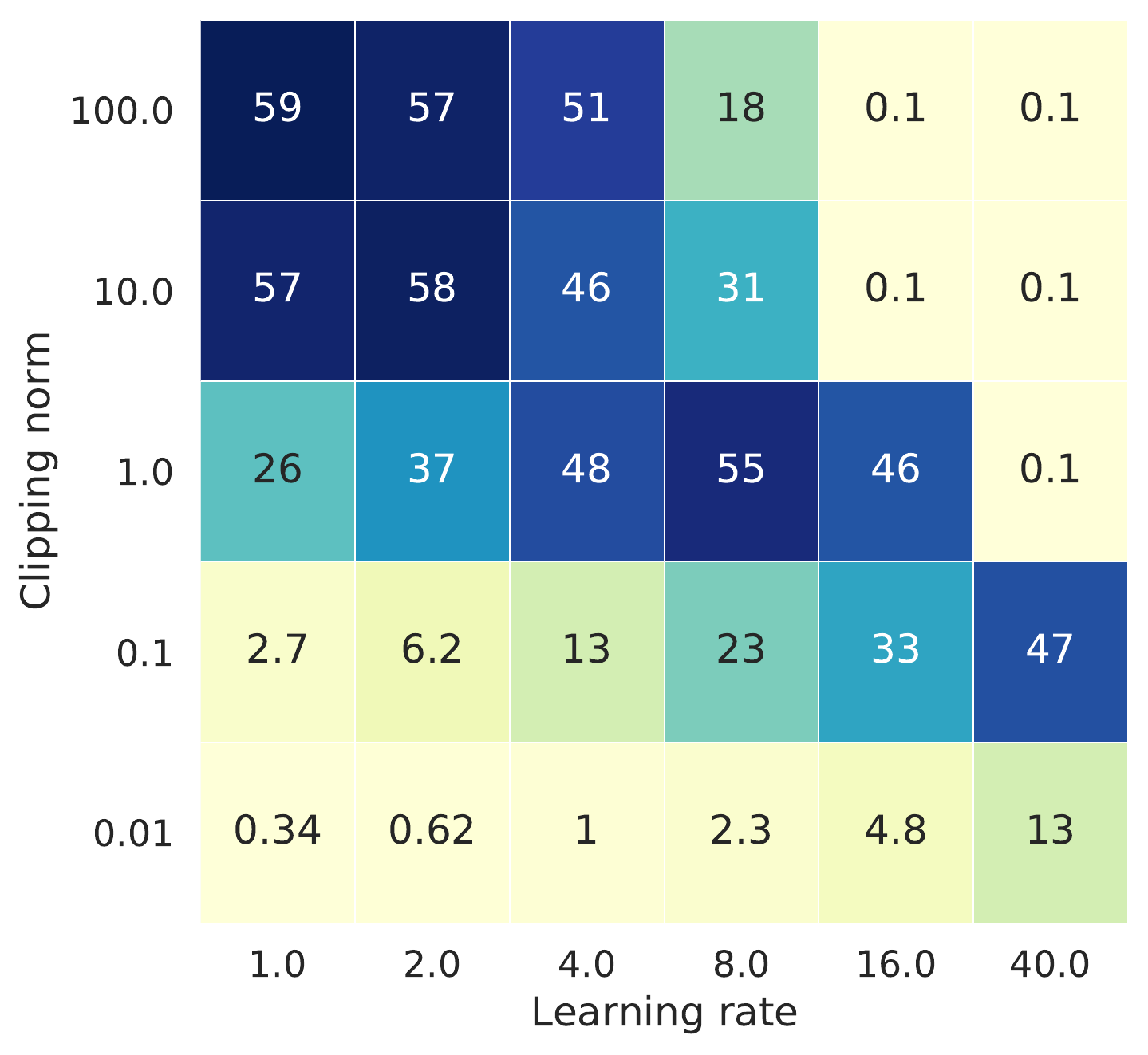}
\caption{Sweep of clipping norm and learning rate when $\sigma = 0$.
Left plot correspond to coarse sweep of hyperparameters,
right plot correspond to more fine-grained sweep of learning rate.
Values in the table is model accuracy.}
\label{fig:c_lr_sweep_sigma0}
\end{figure}

Results for non-private training with gradient clipping are provided in figure~\ref{fig:c_lr_sweep_sigma0}.
As could be seen from the heatmap, model reaches highest accuracy for all $C$ larger than $10$.
We can conclude that when $C > 10$ clipping is no longer happening and it becomes effectively equivalent to training without clipping.

\begin{figure}[p]
\vspace{-1cm}
\begin{subfigure}{.5\textwidth}
  \centering
  \includegraphics[width=\linewidth]{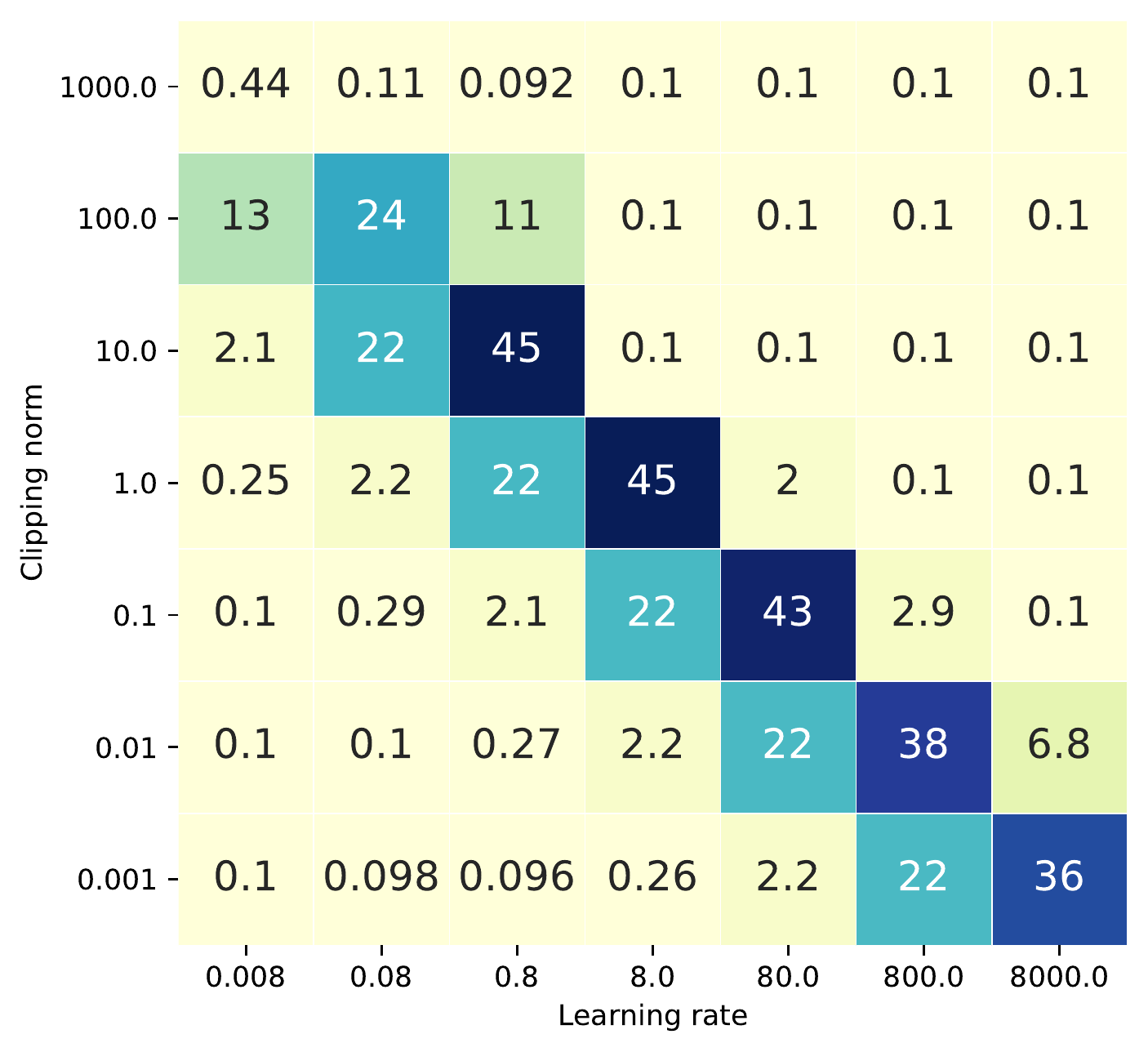}
  \caption{$\sigma / \sqrt{8} = 0.01$}
\end{subfigure}%
\begin{subfigure}{.5\textwidth}
  \centering
  \includegraphics[width=\linewidth]{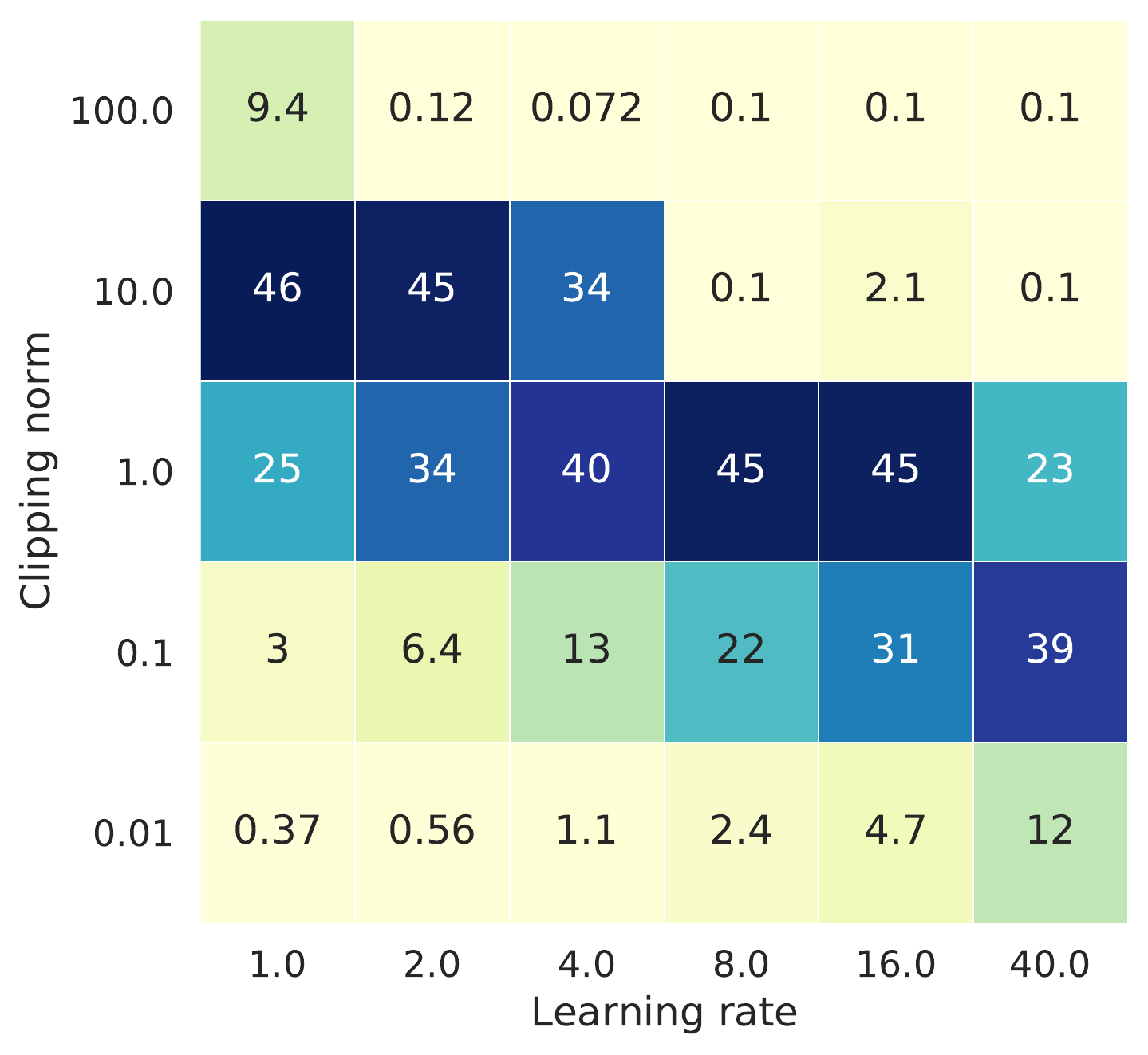}
  \caption{$\sigma / \sqrt{8} = 0.01$}
\end{subfigure}
\begin{subfigure}{.5\textwidth}
  \centering
  \includegraphics[width=\linewidth]{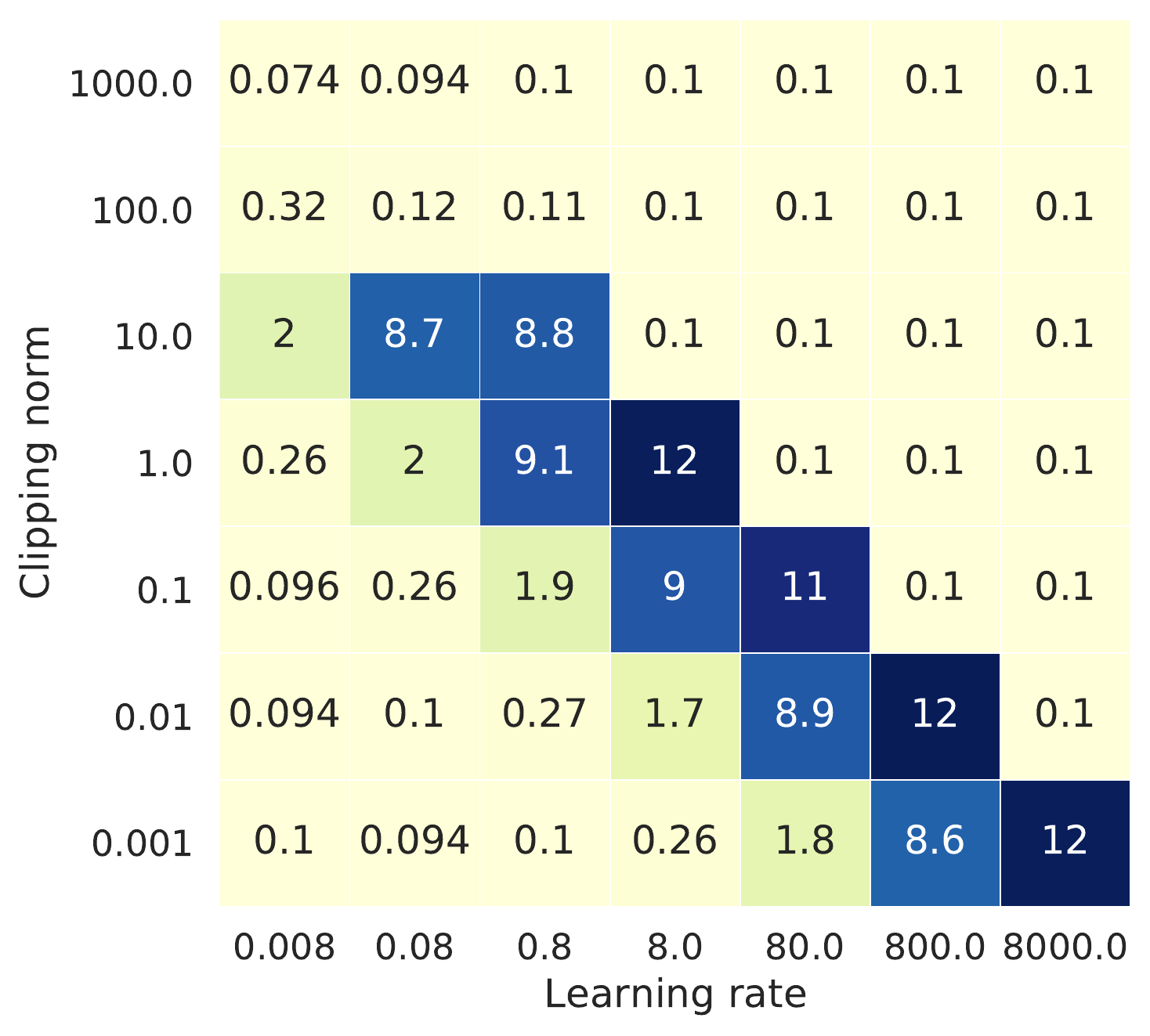}
  \caption{$\sigma / \sqrt{8} = 0.1$}
\end{subfigure}%
\begin{subfigure}{.5\textwidth}
  \centering
  \includegraphics[width=\linewidth]{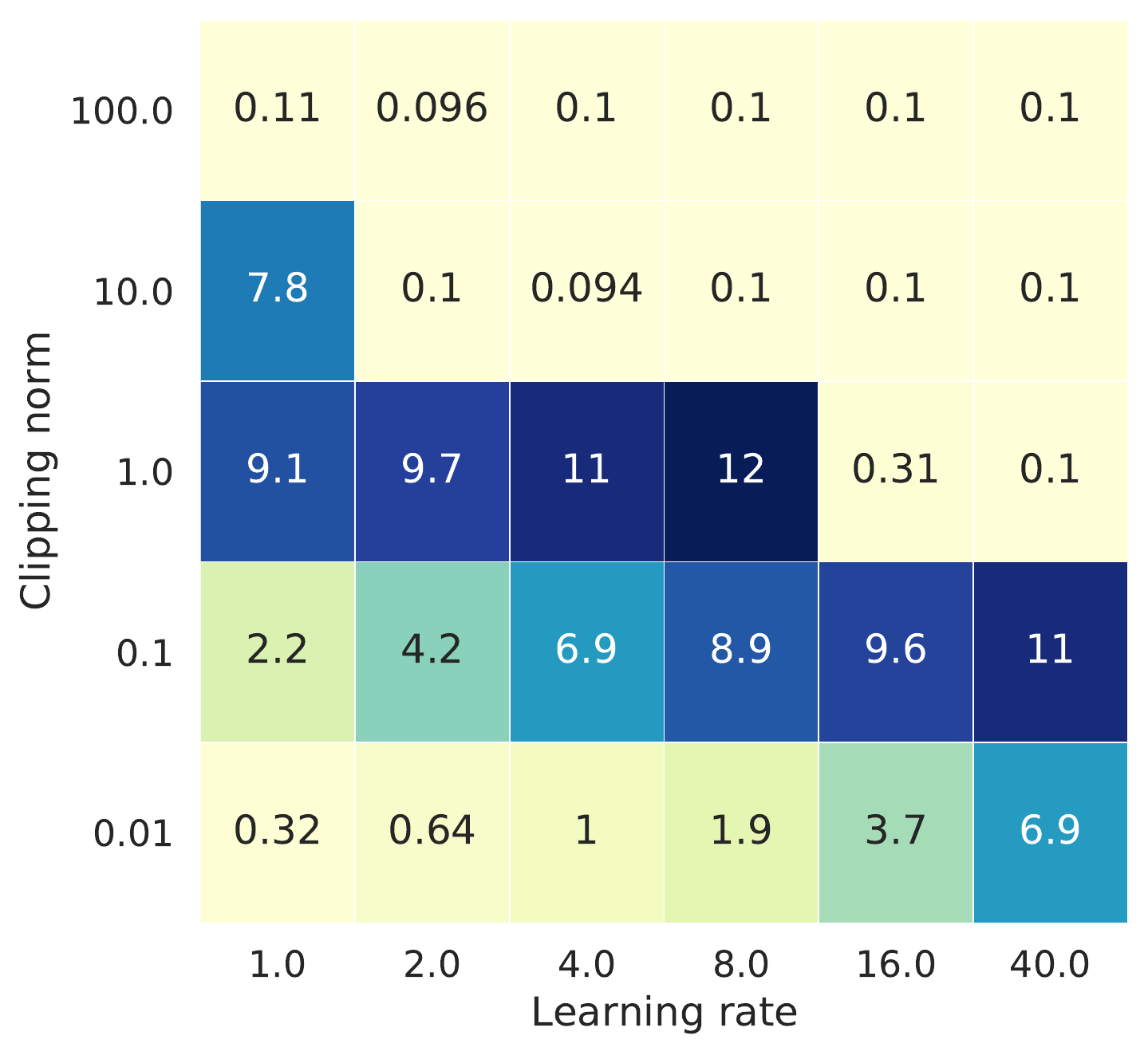}
  \caption{$\sigma / \sqrt{8} = 0.1$}
\end{subfigure}
\begin{subfigure}{.5\textwidth}
  \centering
  \includegraphics[width=\linewidth]{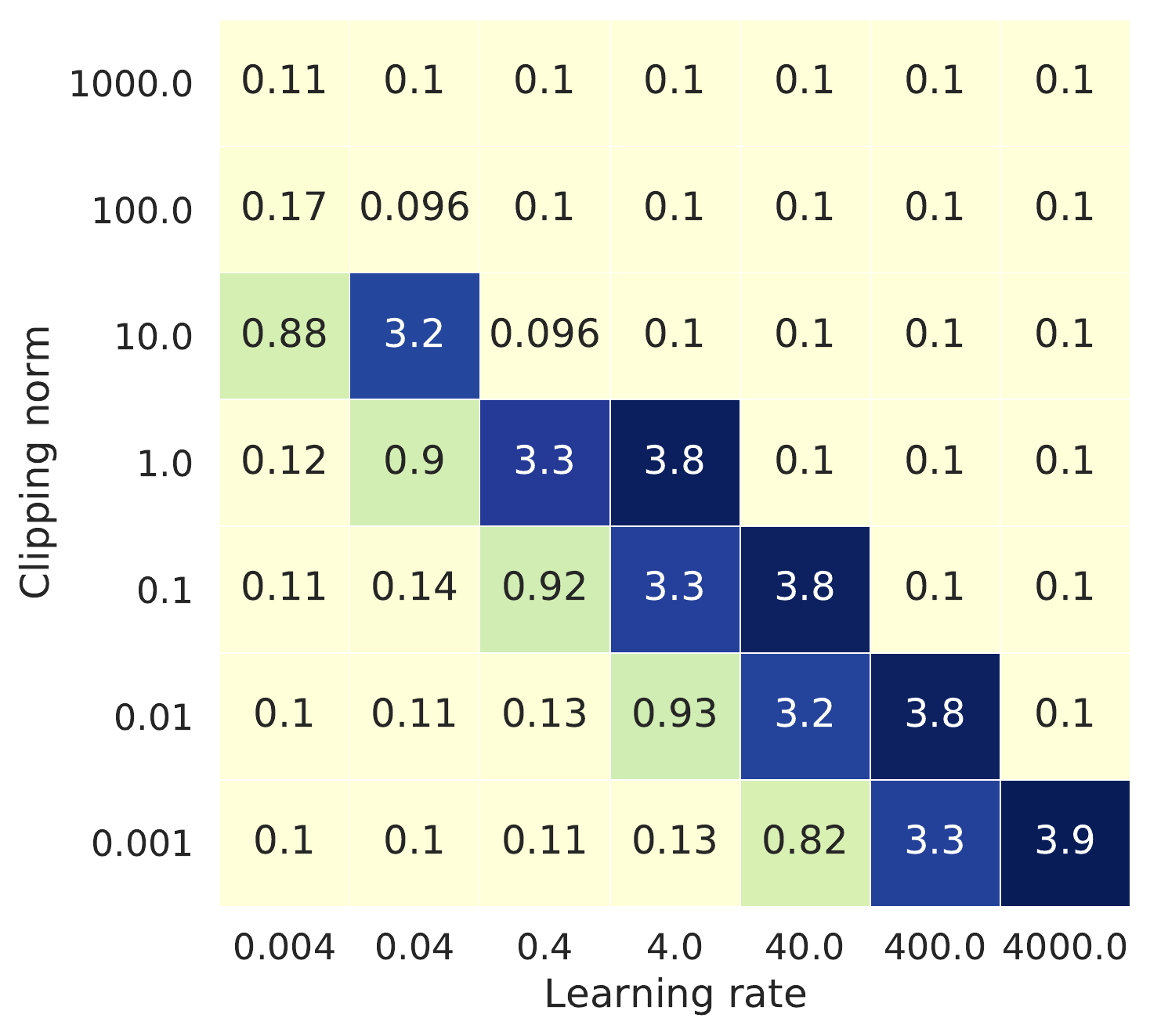}
  \caption{$\sigma / \sqrt{8} = 0.2$}
\end{subfigure}%
\begin{subfigure}{.5\textwidth}
  \centering
  \includegraphics[width=\linewidth]{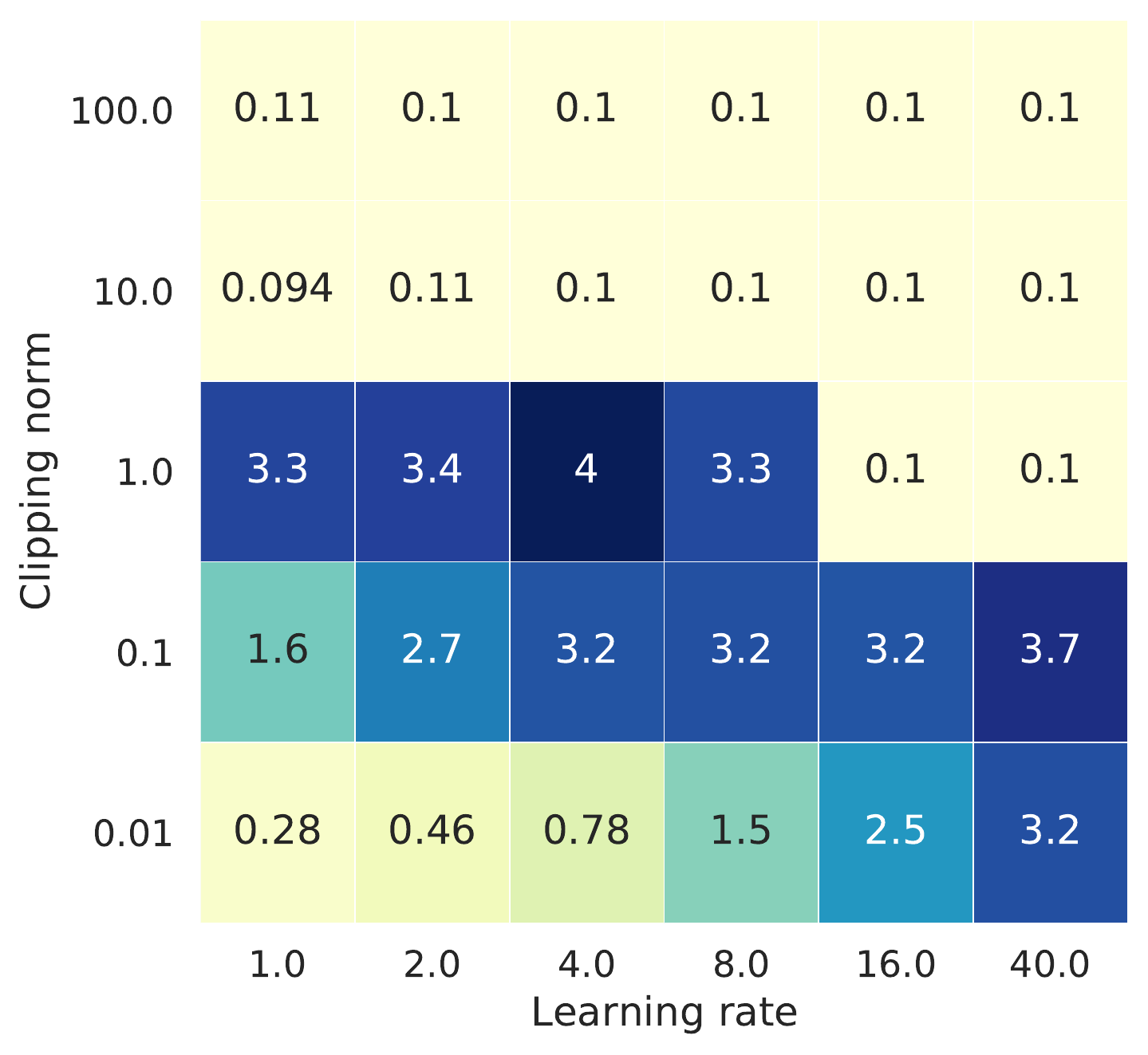}
  \caption{$\sigma / \sqrt{8} = 0.2$}
\end{subfigure}
\caption{Sweep of clipping norm and learning rate with various values of noise.
Left column correspond to coarse sweep of both learning rate and clipping norm,
right column correspond to more fine grained sweeps of learning rate.
Values in the table is model accuracy.}
\label{fig:c_lr_sweep}
\end{figure}

Results with non zero noise multiplier $\sigma$ are provided in figure~\ref{fig:c_lr_sweep}.
In all of these experiments the best accuracy is obtained when $C < 10$. 
This could be explained by the fact that standard deviation of Gaussian noise added to the gradients is computed
as a product of $C$ and $\sigma$.
Thus if $C$ is much large than the norm of the gradients then model updates will be dominated by the noise
which will result in low utility.

Another observation, that the subset of hyperparameters which correspond to the best utility lies
on the curve $\textit{learning\_rate}\cdot C = \textbf{const}$.
This could be explained by the fact that if $C$ is decreased $k$ times then learning rate has to be increased also $k$ times
in order to keep updates roughly the same.
Nevertheless, we observed that extremely high learning rates together with extremely low $C$ typically lead to lower accuracy
and less stable training.

\section{Details of large batch training sweep}\label{app:large_batch}

We explored various combinations of batch size and number of training epochs to study how large batch training can help improving
utility with DP-SGD, see table~\ref{table:imagenet_large_batch}. In these experiments we used the following set of hyperparameters:

\begin{itemize}
    \item Total Gaussian noise added on every training step has standard deviation $\sigma = 0.001 \cdot \sqrt{8} \cdot \frac{\textrm{BatchSize}}{1024}$.
    \item Clipping norm $C$ was set to $1$ in all experiments.
    \item Number of learning rate warmup epochs was set to $\frac{1}{10}$ of total number of training epochs.
    \item Learning rate was set to $16.0$ in all experiments.
    While typically, it's recommended to scale learning rate when increasing the batch size~\cite{goyal2017largebatch,you2017largebatch,hoffer2017largebatch}, it didn't seem to work well in this experiment.
    Specifically we have tried different learning rates from the set $\{4, 8, 16, 32, 64\}$
    with different batch sizes while training model for 10 epochs with DP-SGD.
    We observed that learning rate $16$ was the best or very close to the best for all considered batch sizes.
\end{itemize}

\begin{table}[ht]
\centering
\begin{tabular}{ |c|c|c|c|c|c|c| } 
 \hline
 Batch & & & & & & \\
 size $\rightarrow$ & 1024 & 4*1024 & 16*1024 & 64*1024 & 256*1024 & 1024*1024 \\ \cline{1-1}
 Num &  &  &  &  &  &  \\
 epochs $\downarrow$ &  &  &  &  &  &  \\ 
 \hline
 10 & 56\% & 35.5\% & 5.8\% & 1.6\% & 0.44\% & 0.14\% \\
    & $9.8\cdot 10^8$ & $1.5\cdot 10^7$ & $2.2\cdot 10^5$ & $1.1\cdot 10^3$ & $23$ & $5.6$ \\
 \hline
 20 & 56.4\% & 53.1\% & 18.8\% &  &  &  \\
    & $2\cdot 10^9$ & $3.0\cdot 10^7$ & $4.4\cdot 10^5$ &  &  &  \\
 \hline
 40 & 61.7\% & 57.5\% & 39.5\% & 10\% & 1.3\% & 0.44\% \\
    & $3.9\cdot 10^9$ & $6.1\cdot 10^7$ & $8.9\cdot 10^5$ & $4.2\cdot 10^3$ & 48 & 11.9 \\
 \hline
 80 &  &  & 54.3\% &  &  &  \\
    &  &  & $1.8\cdot 10^6$ &  &  &  \\
 \hline
 160 &  &  & 57.9\% & 37.5\% & 8\% & 0.95\% \\
     &  &  & $3.5\cdot 10^6$ & $1.7\cdot 10^4$ & 120 & 28 \\
 \hline
 640 &  &  &  & 57.2\% & 36.2\% & 6.2\% \\
     &  &  &  & $6.7\cdot 10^4$ & 326 & 72.3 \\
 \hline
\end{tabular}
\caption{ImageNet training using extremely large batches.
Rows correspond to different number of training epochs, columns - to batch size.
Each cell contains two numbers, top one is the accuracy and bottom one is $\epsilon$.
If cell is empty then corresponding experiment was not run.}
\label{table:imagenet_large_batch}
\end{table}

\section{Pre-training on Places365 dataset}\label{app:places365pretraining}

We tried to train Resnet-18 on Places365 for different number of epochs and with different learning rates,
all other hyperparameters were the same as in case of ImageNet training (see appendix A).
Our results are summarized in table~\ref{table:places365accuracy}.
Since accuracy increase going from 40 to 80 epochs was very small we didn't try longer training
and simply picked the training run with the best accuracy as a starting point for all fine-tuning experiments.

\begin{table}[ht]
\centering
\begin{tabular}{ |c|c|c|c|c| } 
 \hline
  Num epochs & 10 & 20 & 40 & 80 \\
 \hline
  Best accuracy & 51.41 & 53.16 & 54.26 & 54.96 \\
 \hline
\end{tabular}
\caption{Accuracy of Resnet-18 trained on Places365 dataset.
Each accuracy number was obtained by sweeping learning rate in $\{0.02, 0.04, 0.08, 0.2, 0.4, 0.8, 1.6\}$.
}
\label{table:places365accuracy}
\end{table}

\section{DP-SGD finetuning experiments}\label{app:finetuning}

\begin{figure}[ht]
\centering
\includegraphics[width=\textwidth]{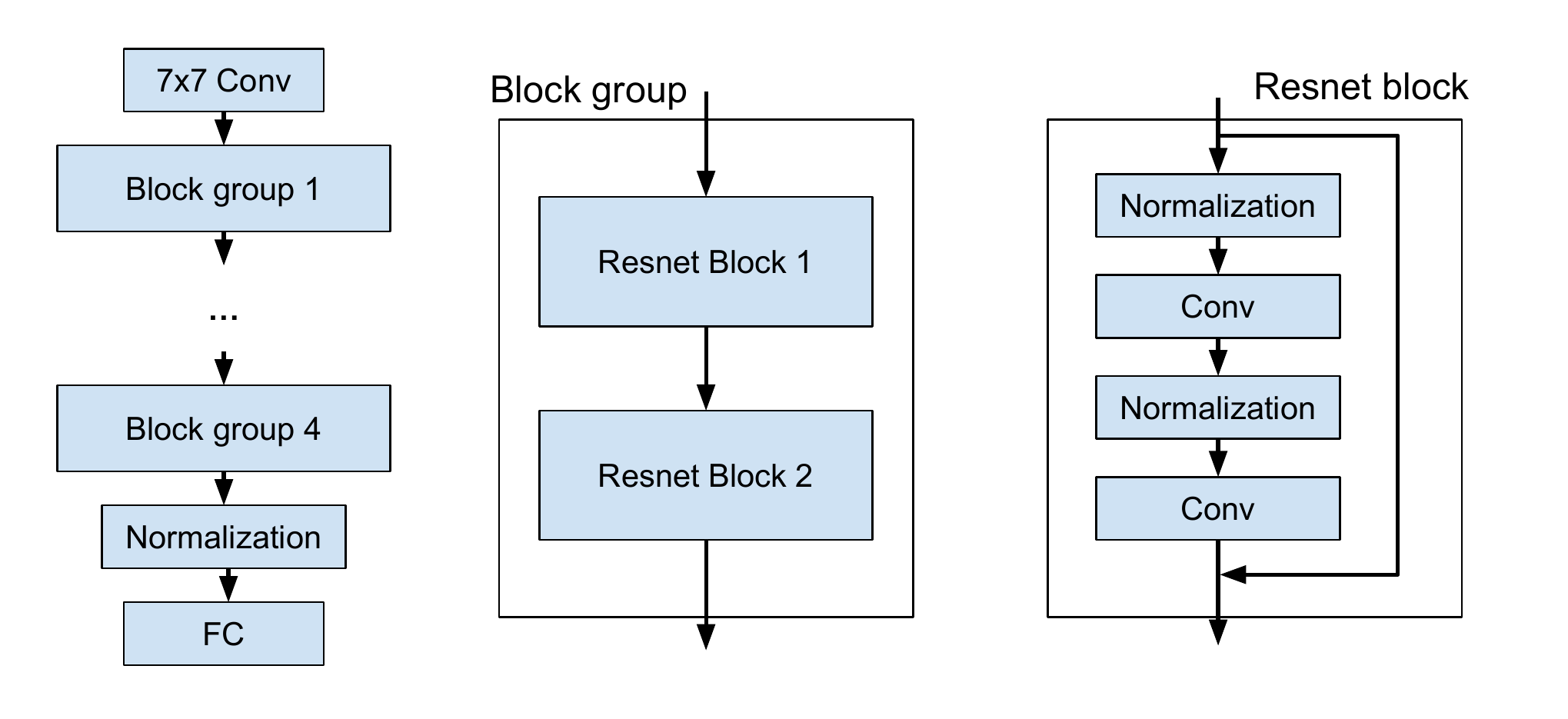}
\caption{Structure of Resnet-v2-18.
Resnet18 contains 4 block groups.
There is an input 7x7 convolution before the first block group and fully connected logits layer after the last block group.
Each block group contains 2 residual blocks.
Each residual blocks contains two convolutions with normalization layers and residual connection.
Note that some of the residual blocks had projection and/or pooling operation on their residual connection.}
\label{fig:resnet18structure}
\end{figure}

We performed finetuning experiments on Resnet18 while optionally freezing some of the layers.
Due to the structure of Resnet18 (see figure~\ref{fig:resnet18structure}) we decided that it's most convenient
to freeze layers at a {\it block group} level.
So we did experiments for the following configurations of frozen/trainable layers:

\begin{itemize}
    \item \textbf{None frozen block groups.} Entire network is trainable.
    \item \textbf{1 frozen block groups.} Input convolution and block group 1 are frozen.
    \item \textbf{2 frozen block groups.} Input convolution, block groups 1 and 2 are frozen.
    \item \textbf{3 frozen block groups.} Input convolution, block groups 1, 2 and 3 are frozen.
    \item \textbf{4 frozen block groups.} Everything except the last fully connected layer is frozen.
      Only last fully connected layer (logits layer) is trainable.
\end{itemize}

\begin{table}[ht]
\centering
\begin{tabular}{ |c|c|c|c|c|c| } 
 \hline
  Frozen block groups & None & 1 & 2 & 3 & 4  \\
 \hline
  Best accuracy & $1.8\%$ & $2.1\%$ & $2.7\%$ & $3.3\%$ & $23.2\%$ \\
 \hline
\end{tabular}
\caption{Private finetuning experiments with frozen layers. Training was done for 10 epochs with privacy budget $\epsilon \approx 6$.
In this experiments we used Nesterov momentum optimizer with cosine learning rate decay and learning rate warmup for 1 epoch.
We used batch size $1024*1024$.
Each accuracy number was obtained by sweeping learning rate in $\{4.096, 12.288, 40.96, 122.88, 409.6, 1228.8, 4096.0\}$}
\label{table:finetuning1}
\end{table}

In the first set of experiments we did initial learning rate sweep using batch size 1024 to identify feasible range of learning rates for finetuning,
and then run a series of DP-SGD experiments with frozen layers and batch size 1024*1024, see table~\ref{table:finetuning1}.
As could be seen from the table, when number of frozen block groups is less than 4 accuracy stays pretty low.
Thus we focused further experiments on cases of 4 or 3 frozen block groups.

\begin{table}[ht]
\centering
\begin{tabular}{ |c|c|c| } 
 \hline
  Frozen block groups & 3 & 4  \\
 \hline
  Best non private accuracy & $66.7\%$ & $41.7\%$ \\
 \hline
\end{tabular}
\caption{Best results of non-private finetuning of Resnet18 model.
All experiment were run for 80 epochs with batch size 1024
and by sweeping learning rate in $\{0.004, 0.012, 0.04, 0.12, 0.4, 1.2, 4.0\}$.}
\label{table:non_private_finetuning}
\end{table}

For 3 and 4 frozen block groups we did non private finetuning to estimate upper bound of accuracy
which is possible to achieve, see table~\ref{table:non_private_finetuning}.

\newcommand{\simpleVGG}[1]{simpleVGG{#1}\xspace}

\section{Experiments on CIFAR-10}\label{app:cifar10}

We aim to examine the effect of model size on the privacy-utility tradeoff. We consider a class of neural networks which includes the one used for differentially private CIFAR-10 classification in a few previous work~\cite{papernot2020tempered, kairouz2021practical}.
This class of networks, which we denote as \simpleVGG{}, can be abstracted in the following way. 
A \simpleVGG{} consists of multiple ``blocks'' and a linear layer, connected by max-pooling layers; each ``block'' consists of multiple convolution layers with the same number of channels. 
In the experiments, we will consider three sets of \simpleVGG{}, in which we vary the number of convolution channels, number of layers per block, and the number of blocks, respectively. 
We fix the fully connected layer to size 128, and use tanh as the activation function. We use \simpleVGG{-32(2)-64(2)-128(2)-128} to denote a network with three blocks, each having two convolution layers with 32, 64 and 128 channels, and a fully connected layer of size 128 in the end. This is the neural network that has been used in previous work~\cite{papernot2020tempered, kairouz2021practical}.

\begin{figure}[ht]
\centering
\includegraphics[width=\linewidth]{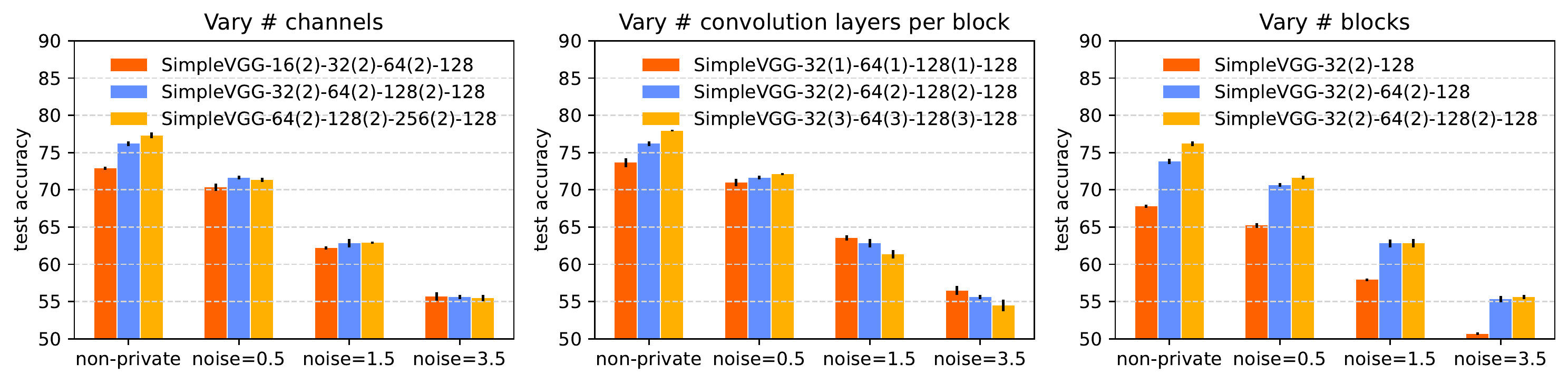}
\caption{Test accuracy for \simpleVGG{s} on CIFAR-10. Averaged over $3$ repeated runs.}
\label{fig:cifar10_bar}
\end{figure}

We train each neural network on CIFAR-10 for $100$ epochs with batch size $500$. We consider three different privacy levels for each setting: noise $\sigma=0.5$, $1.5$ and $\sigma=3.5$, corresponding to $\epsilon=47.41$, $3.45$, and $1.20$ at $\delta=10^{-5}$. 
For each setting, we keep the clip norm to be $1.0$ and use a momentum optimizer with learning rate tuned from $\{0.001, 0.002, 0.005,\dots, 0.1, 0.2, 0.5\}$.
In Figure~\ref{fig:cifar10_bar}, we plot the final test accuracy of 7 different \simpleVGG{s} under different privacy levels, each with the learning rate that achieves the best final test accuracy. We compare networks with different number of convolution channels, different number of convolution layers per block, and different number of blocks in the three subplots. Clearly, in the non-private setting, accuracy increases with the complexity of the network. As $\sigma$ grows, the gap becomes smaller and the simpler network gets the best accuracy when we vary the number of channels and number of convolution layers per block.

\section{Values of \texorpdfstring{$\epsilon$}{epsilon} at different \texorpdfstring{$\delta$}{delta}.}\label{app:eps_detla_tradeoff}

In most of the text of the paper we provide $(\epsilon, \delta)$-DP guarantees
for fixed $\delta = 10^{-6}$, i.e. $\delta \approx \frac{1}{\textrm{DATASET SIZE}}$.
In practice, these privacy guarantees come from DP-SGD privacy accountant and could
be recomputed for different values of $\delta$.

\begin{figure}[ht]
\centering
\includegraphics[width=\textwidth]{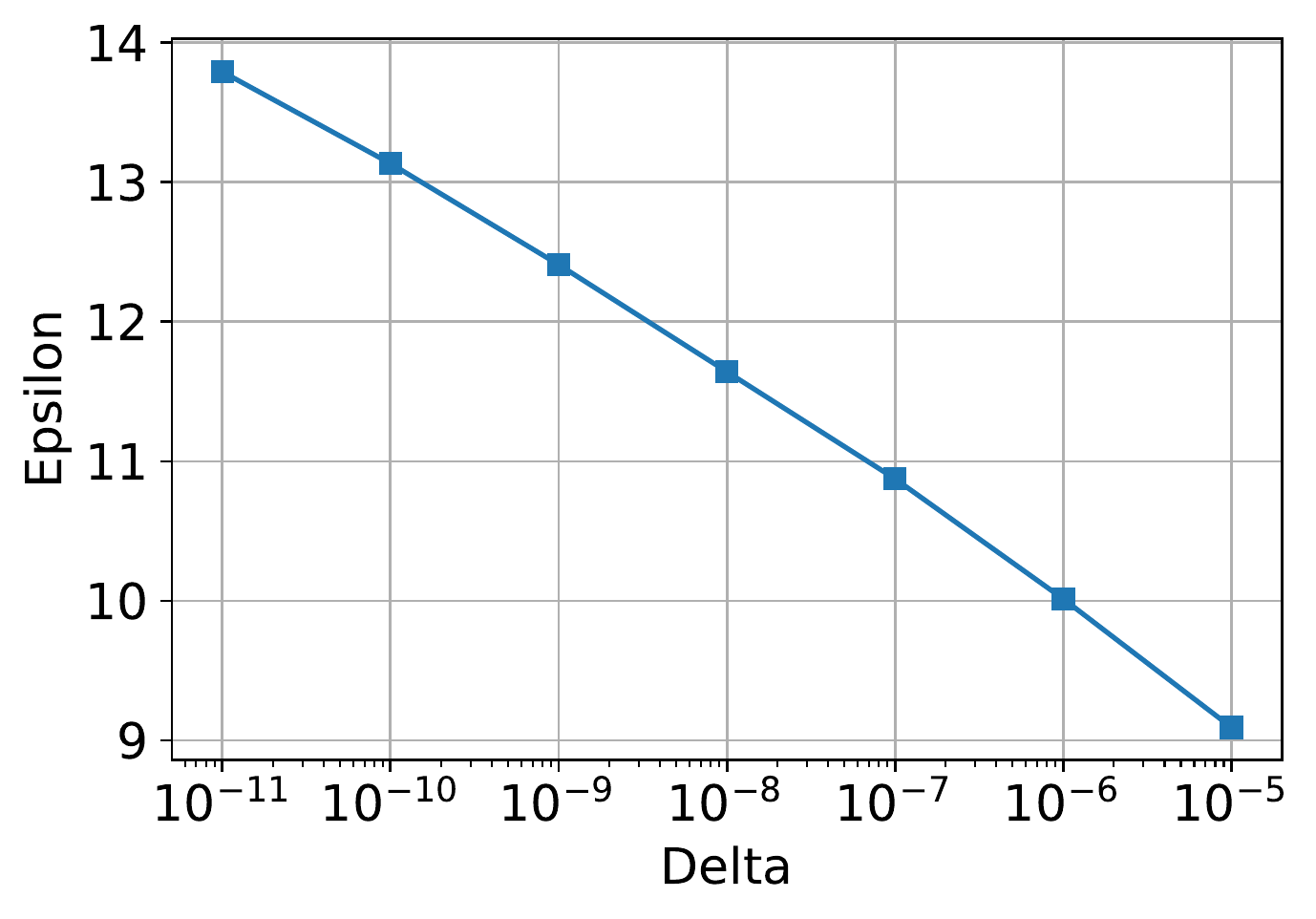}
\caption{Relationship between $\epsilon$ and $\delta$ for our best training run,
which achieves  $\approx 48\%$ top-1 accuracy.}
\label{fig:eps_delta_best_run}
\end{figure}

Figure~\ref{fig:eps_delta_best_run} shows $\epsilon$ computed at different values of $\delta$
for our best training run, which achieves $\approx 48\%$ final top-1 accuracy.

\end{document}